\documentclass{article}
\usepackage{ijcai17}
\usepackage{amsmath}
\usepackage{amsfonts}
\usepackage{amssymb}
\usepackage{graphicx}
\usepackage[font=small]{caption}
\usepackage{subfig}

\usepackage{hyperref}
\usepackage{enumitem}

\usepackage{times}
\usepackage{xparse}

\usepackage{txfonts}

\usepackage[bbgreekl]{mathbbol}

\usepackage[usenames,dvipsnames,svgnames,table]{xcolor}
\usepackage{tikz}
\usetikzlibrary{calc,tikzmark}
\usetikzlibrary{fit}
\usetikzlibrary{positioning}
\usetikzlibrary{decorations.markings}
\usetikzlibrary{shapes.geometric}
\usetikzlibrary{shapes.multipart}
\usetikzlibrary{bayesnet}
\definecolor{Red}{rgb}{1,0,0}
\definecolor{Green}{rgb}{0,0.69,0}
\definecolor{Blue}{rgb}{0,0,1}
\definecolor{LightBlue}{rgb}{0,0.5,1}
\definecolor{veryLightBlue}{rgb}{0.85,0.98,1}
\definecolor{veryLightGreen}{rgb}{0.6,1,0.6}
\definecolor{Skin}{rgb}{1,0.71,0.69}
\definecolor{Grey}{rgb}{0.5,0.5,0.5}
\definecolor{LightGrey}{rgb}{0.6,0.6,0.6}
\definecolor{Black}{rgb}{0,0,0}
\definecolor{White}{rgb}{1,1,1}
\newcommand{\red}{\color{Red}}
\newcommand{\green}{\color{Green}}
\newcommand{\blue}{\color{Blue}}

\newcommand{\orange}{\color{Bittersweet}}

\newcommand{\violet}{\color{BlueViolet}}

\usepackage{microtype}
\usepackage[compact]{titlesec}
\hypersetup{
    colorlinks,
    linkcolor={black},
    citecolor={black},
    urlcolor={black}
}

\usepackage[disable]{todonotes} 
\usepackage{verbatim}

\NewDocumentCommand\placeholder{mmo}{
  \framebox{
    \begin{minipage}[c][#2]{#1}
      \centering
      \IfNoValueTF{#3}{Placeholder}{#3}
    \end{minipage}
  }
}

\usepackage{mathtools}
\usepackage{etoolbox}

\newcommand*{\ctelementsep}{}%
\newcommand*{\ctlastelement}{}%
\newcommand*{\ctprelastelement}{}%

\newcommand*{\ctdodisplayelement}[1]{%
  \ctelementsep
  \ctlastelement
  \renewcommand{\ctlastelement}{%
    \renewcommand{\ctelementsep}{, }%
    \renewcommand{\ctprelastelement}{, and }%
    \citeauthor{#1} [\citeyear{#1}]%
  }}%

\newcommand*{\citet}[1]{%
  \renewcommand*{\ctelementsep}{}%
  \renewcommand*{\ctlastelement}{}%
  \renewcommand*{\ctprelastelement}{}%
  \forcsvlist{\ctdodisplayelement}{#1}%
  \ctprelastelement \ctlastelement
}

\title{Temporal Grounding Graphs for Language Understanding with Accrued Visual-Linguistic Context}

\author{Rohan Paul \textmd{and} Andrei Barbu \textmd{and} Sue Felshin \textmd{and} 
Boris Katz \textmd{and} Nicholas Roy\thanks{R. Paul and A. Barbu 
contributed equally to this work. R. Paul and N. Roy are with the Robust Robotics 
Group and A. Barbu, S. Felshin and B. Katz are with the Center for 
Brains, Minds, and Machines and the Computer Science and AI Lab (CSAIL) at MIT.\@
Contact: \{rohanp, abarbu, sfelshin, boris, nickroy\}@csail.mit.edu} \\
Massachusetts Institute of Technology, Cambridge, MA
}

\begin{document}
\maketitle

\begin{abstract}
A robot's ability to understand or \emph{ground} natural language instructions is fundamentally
tied to its knowledge about the surrounding world.
We present an approach to grounding natural language utterances in the context
of factual information gathered through natural-language interactions and past
visual observations.
A probabilistic model estimates, from a natural language utterance, the objects,
relations, and actions that the utterance refers to, the objectives for future
robotic actions it implies, and generates a plan to execute those actions while
updating a state representation to include newly acquired knowledge from the
visual-linguistic context.
Grounding a command necessitates a representation for past observations and
interactions; however, maintaining the full context consisting of all possible
observed objects, attributes, spatial relations, actions, etc., over time is
intractable.
Instead, our model, \emph{Temporal Grounding Graphs}, maintains a learned
state representation for a belief over factual groundings, those
derived from natural-language interactions, and lazily infers new groundings
from visual observations using the context implied by the utterance.
This work significantly expands the range of language that a robot can
understand by incorporating factual knowledge and observations of its workspace
in its inference about the meaning and grounding of natural-language
utterances.

\end{abstract}


\NewDocumentCommand\Trajectory{}{ \mu_{t}(t) }

\NewDocumentCommand\Control{}{ \mu_{t} }
\NewDocumentCommand\ControlInitial{}{ \mu_{t}(t'_{i}) }
\NewDocumentCommand\ControlFinal{}{ \mu_{t}(t'_{f}) }

\NewDocumentCommand\Action{m}{%
        \IfNoValueTF{#1}
        { \mu }
        { \mu_{#1} }
}


\NewDocumentCommand\KB{m}{%
        \IfNoValueTF{#1}
        { K }
        { K_{#1} }
}

\NewDocumentCommand\PersistentWorldModel{m}{%
        \IfNoValueTF{#1}
        { \omega }
        { \omega_{#1} }
}

\NewDocumentCommand\Language{m}{%
        \IfNoValueTF{#1}
        { \Lambda }
        { \Lambda_{#1} }
}

\NewDocumentCommand\Grounding{m}{%
        \IfNoValueTF{#1}
        { \Gamma }
        { \Gamma_{#1} }
}

\NewDocumentCommand\GV{m}{%
        \IfNoValueTF{#1}
        { \gamma }
        { \gamma_{#1} }
}

\NewDocumentCommand\Phrase{m}{%
        \IfNoValueTF{#1}
        { \lambda }
        { \lambda_{#1} }
}

\NewDocumentCommand\Images{m}{%
        \IfNoValueTF{#1}
        {\mathcal{I}}
        {\mathcal{I}_{#1}}
}

\NewDocumentCommand\Image{m}{%
        \IfNoValueTF{#1}
        {I}
        {I_{#1}}
}

\NewDocumentCommand\VisualObservations{m}{%
        \IfNoValueTF{#1}
        {Z}
        {Z_{#1}}
}

\NewDocumentCommand\VisualObservation{oo}{%
    \IfNoValueTF{#2}
    {
        \IfNoValueTF{#1}
        {z}
        {z^{#1}}
    }
    { z^{#1}_{#2} }
}

\NewDocumentCommand\Correspondences{m}{%
        \IfNoValueTF{#1}
        {\Phi}
        {\Phi_{#1}}
}

\NewDocumentCommand\CV{m}{%
        \IfNoValueTF{#1}
        {\phi}
        {\phi_{#1}}
}

\NewDocumentCommand\WorldModel{m}{%
        \IfNoValueTF{#1}
        {\Upsilon}
        {\Upsilon_{#1}}
}

\NewDocumentCommand\TrackerFullState{m}{%
        \IfNoValueTF{#1}
        {\mathbb{S}}
        {\mathbb{S}_{#1}}
}

\NewDocumentCommand\Tracker{oo}{%
    \IfNoValueTF{#2}
    {
        \IfNoValueTF{#1}
        {S}
        {S^{#1}}
    }
    { S^{#1}_{#2} }
}

\NewDocumentCommand\GroundingFullState{m}{%
        \IfNoValueTF{#1}
        {\mathcal{G}}
        {\mathcal{G}_{#1}}
}

\NewDocumentCommand\GroundingFactored{oo}{%
    \IfNoValueTF{#2}
    {
        \IfNoValueTF{#1}
        {\Gamma}
        {\Gamma^{#1}}
    }
    { \Gamma^{#1}_{#2} }
}

\NewDocumentCommand\Dec{m}{ #1^{\delta} }

\NewDocumentCommand\Imp{m}{ #1^{\pi} }

\NewDocumentCommand\Per{m}{ #1^{\mathcal{P}} }

\NewDocumentCommand\Fact{m}{ #1^{\mathcal{F}} }

\NewDocumentCommand\Belief{m}{ \emph{bel}(#1) }

\NewDocumentCommand\Integral{m}{ \int_{#1} }

\NewDocumentCommand\Sum{m}{ \sum_{#1} }

\NewDocumentCommand\Size{m}{ \lvert {#1} \rvert }

\NewDocumentCommand\NumTrackers{}{ n_{\mathcal{O}} }
\NewDocumentCommand\NumDetections{}{ n_{t} }
\NewDocumentCommand\NumDecGroundings{}{ n_{\delta} }

\NewDocumentCommand\Predicates{}{ \Omega }

\NewDocumentCommand\GroundingType{}{ \sigma }

\NewDocumentCommand\Role{}{ \mathcal{R} }

\NewDocumentCommand\SymWM{}{ \Upsilon }

\NewDocumentCommand\Time{m}{ #1_{t} }

\NewDocumentCommand\Context{}{ \{\Language{0:t-1}, \VisualObservations{0:t-1} \} }

\section{Introduction}
Effective human-robot interaction in homes or other complex dynamic workspaces
requires a linguistic interface; robots should understand what owners want them
to do.
This is only possible with a shared representation of the environment, both past
and present, as well as a mutual understanding of the knowledge exchanged in
prior linguistic interactions.
At present, humans and robots possess disparate world representations which do
not lend themselves to enabling natural human-robot interactions.
Humans possess a continuously-expanding rich understanding of the environment
consisting of semantic entities and higher-order relationships which includes
knowledge about previous interactions and events.
In contrast, robots estimate a metric picture of the world from their
sensors which determine which action sequences to take.
%
%
Our goal is to bridge this \emph{semantic gap} and enable robots, like humans,
to acquire higher-order semantic knowledge about the environment through
experience and use that knowledge to reason and act in the world.
To this end, we develop an approach to building robotic systems which follow
commands in the context of accrued visual observations of their workspace and
prior natural-language interactions.
Natural-language interactions can include both commands to perform an action or
important factual information about the environment.
The robot understands and retains facts for later use and is able to execute the
intended tasks by reasoning over acquired knowledge from the past.

In recent years, probabilistic models have emerged that interpret or \emph{ground} natural language in the context of a 
robot's world model. 
%
%
Approaches such as those of 
\citet{tellex2011understanding,howard2014efficient,chung2015performance,Paul-RSS-16}, 
relate input language to entities in the world and actions to be performed by
the robot by structuring their inference according to the parse or semantic
structure embedded in language.
A related set of approaches such as those of
\citet{zettlemoyer2007online,chen2010training,kim2012unsupervised,artzi2013weakly} employ
semantic parsing to convert a command to an intermediate representation, either
$\lambda$-calculus or a closely related logic formalism, which can then be executed by
a robot.
A key limitation of current models is their inability to reason about or reference past observations. 
In essence, many models assume that the world is static, facts do not change,
and that perception is entirely reliable.
\citet{MatuszekAAAI2014} present a formulation that allows uncertainty in the
knowledge about object in the scene and present an approach that learns 
object attributes and spatial relations through joint reasoning with
perceptual features and language information.
In a similar vein,
\citet{chen2011learning,guadarrama2013grounding,walter14a,hemachandra15,andreas2015alignment},
demonstrate systems more resilient in uncertainty in the input given the need
for the robot to carry out actions.
Even a simple statement such as ``The fruit I placed on the table is my
snack'' followed by the command ``Pack up my snack'' requires reasoning about
previous observations of the actions of an agent.
While work such as that of \citet{misra2016tell, liu2016taskstructure}
incorporates context it does so by learning the static relationships between
properties of the environment and actions to be performed; a different task from
learning facts online and remembering the actions of agents.

Complementary efforts in many communities have focused on acquiring rich
semantic representations from observations or knowledge-based systems.
These include efforts in the vision community towards compositional activity
recognition~\cite{barbu2013saying,yu2015compositional} and work such as that by
\citet{berzak2016you} which combines activity recognition with language
disambiguation.
Further, approaches for associating language with semantic constructs have been
explored in the semantic parsing community~\cite{berant2013freebase}.
\citet{cantrell2010robust} investigate a closely-related problem, grounding
robotic commands which contain disfluencies to objects in images.
We consider a broader notion of grounding than \citet{cantrell2010robust} in
which we include the actions of agents as well as knowledge from prior
linguistic interactions but note that their approach is complementary and could
be used to increase the robustness of the work presented here.

We seek the ability to reason about the future actions of a robot using
knowledge from past visual observations and linguistic interactions.
Enabling such reasoning entails determining \emph{which} symbols grounded by 
past observations should be retained to enable future inferences. 
One approach is to estimate and propagate all symbols from past percepts.
This is intractable as the space of all static or dynamic semantic
relations between all observed objects is exponentially large and continues to grow 
exponentially as new relations are learned.
Alternatively, one can forgo symbol propagation and retain raw observations alone.
This approach incurs a linear storage cost, but requires jointly interpreting
the current utterance with all past utterances, combinatorially increasing
the inference cost with each utterance.

In this work, we present \emph{Temporal Grounding Graphs}, a probabilistic model that 
enables incremental grounding of natural language utterances using learned knowledge from accrued 
visual observations and language utterances.
%
The model allows efficient inference over the constraints for future actions 
in the context of a rich space
of perceptual concepts such as object class, shape, color, relative position, mutual actions,
etc. as well as factual concepts (does an object belong to an agent such as a human) that grow over time. 
%
Crucially, the approach attempts to balance the computational cost of incremental inference versus 
the space complexity of knowledge persistence.
%
Our model maintains a learned representation as a belief over 
factual relations from past language. 
The model accrues visual observations but delays estimation of grounding of perceptual concepts. 
Online, the model estimates the necessary past visual context required for interpreting the instruction. 
The context guides the construction of a constrained 
grounding model that performs focused time-series inference over past visual observations. 
This approach incurs an additional inference cost online
but reduces the need for exhaustively estimating all perceptual groundings from past observations.
Factors in the model are trained in a data-driven manner using an aligned vision-language corpus.
We demonstrate the approach on a Baxter Research Robot following and executing 
complex natural language instructions in a manipulation domain using a standardized object data set.

\section{Problem Formulation}

We consider a robot manipulator capable of executing a control sequence 
$\Control$ composed of an end-effector trajectory $\{ \ControlInitial, \dots, \ControlFinal \}$ initiated at time $t$. 
The robot's workspace consists of objects $\mathcal{O}$ 
that may be 
manipulated by the robot or other agents such as the human operator.
We assume that each object possesses a unique identifier (a symbol) and an initial metric pose known 
\emph{a-priori} to the robot. 
The robot observes its workspace through a visual sensor that collects  
an image $\Images{t}$ at time $t$,  each associated with metric depth information for localization within the environment. 
Further, we assume the presence of an object recognition system for a known set of object classes which 
yields potential detections (each a sub-image) with an associated class likelihood. 
Let $\VisualObservations{t}$ denote the set of detections 
$ \{ \VisualObservation[t][0], \dots,\VisualObservation[t][n_{t}] \}$ from $\Image{t}$ 
where $n_{t}$ is the total number of detections. 
The human operator communicates with the robot through a natural language interface 
(via speech or text input) either instructing the robot to perform actions or providing 
factual information about the workspace. 
Let $\Language{t}$ denote the input language utterance from the human received at time $t$ 
which can be decomposed into an ordered tree-structured set of phrases 
$\{\Phrase{1}, \dots, \Phrase{n} \}$ using a parsing formalism.
We consider the problem of enabling the robot to understand the language 
utterance from the human in the context of learned workspace knowledge 
from past visual observations or factual information provided by the human.
%
Next, we define the space of concepts that represent \emph{meaning}
conveyed in a language utterance and subsequently formalize the grounding problem. 
\subsection{Grounding Symbols} 
We define a space of concepts that characterize the semantic knowledge about the robot's workspace. 
The robot's workspace can be expressed as a set of symbols associated with semantic entities 
present in the environment. 
These include object instances, the human operator and the robot itself, and form the 
symbolic world model $\SymWM$. 
Concepts convey knowledge about the properties or attributes associated with entities or relationships
between sets of entities. 
It is common to employ a symbolic predicate-role representation for concepts~\cite{russell1995modern}. 
A predicate expresses a relation $\GroundingType \in \Predicates$ defined over a set of symbols $\Role$, 
each associated with an entity in the world model $\SymWM$. 
The space of possible predicates that may be true for a world representation forms the space of grounding symbols  
$\Grounding{}$: 
\begin{equation}
\Gamma = 
\left\{ 
  \hspace{1mm}
  \gamma^{\GroundingType}_{ \Role}
  \hspace{1mm} | \hspace{1mm} 
  \GroundingType \in \Predicates, 
  \Role \subseteq \SymWM
  \hspace{1mm}
\right\}.
\end{equation}
For example, if the workspace model consists of a human and a box denoted as symbols $o_{1}$ and $o_{2}$,  
the grounding for the event that involves a person lifting the box object is represented by the grounding 
symbol $\textrm{PickUp}(o_{1},o_{2})$.%
%
\begin{table} 
\begin{center}
  \vspace{2ex}\caption[h]{\footnotesize{The space of grounding symbols and their arities. Object classes
    are provided and learned by object detectors. Regions, modifiers, human
    actions and planner constraints are learned as described in Section~\ref{subsec:training}.
    The space of factual concepts is open and expands through natural-language
    interaction.}}
\begin{tabular}{ l | l }\label{tab:gnd-symbols}%
    Agents & \footnotesize{\textrm{Robot$_1$}, \textrm{Human$_1$}} \\
    Objects & \footnotesize{\textrm{Block$_1$}, \textrm{Can$_1$}, \textrm{Box$_1$}, \textrm{Fruit$_1$}, \textrm{Cup$_1$}, \ldots}\\
    Regions & \footnotesize{\textrm{LeftOf$_2$}, \textrm{InFrontOf$_2$}, \textrm{OnTopOf$_2$}, \ldots} \\
    Modifiers & \footnotesize{\textrm{Quickly$_1$}, \textrm{Slowly$_1$}, \textrm{Big$_2$}, \textrm{Red$_2$}, \ldots}\\
    Human actions & \footnotesize{\textrm{PickUp$_2$}, \textrm{PutDown$_2$}, \textrm{Approach$_2$}, \ldots} \\
    Factual concepts & \footnotesize{\textrm{Mine$_1$}, \textrm{Favourite$_1$}, \textrm{Forbidden$_1$}, \ldots} \\
    Planner constraints & \footnotesize{\textrm{Intersect$_2$}, \textrm{Contact$_2$}, \textrm{SpatialRelation$_2$}, \ldots} \\
  \end{tabular}
\end{center}
\vspace{-1ex}
\end{table} 

%
Grounding symbols can be categorized in terms of the type of concepts they convey about the workspace. 
A set of \emph{declarative} grounding symbols  $\Dec{ \Grounding{} }$ 
convey knowledge about the robot's workspace.  
These include perceptual concepts  $\Per{ \Grounding{} } $ consisting of
entities such as objects and agents,
static relations like spatial regions (e.g, on, left, front) and
event relations that denote interactions (e.g., place, approach, slide)
between agents and objects present in the scene.
Further, declarative groundings include \emph{factual} concepts  $\Fact{ \Grounding{} }$
consisting of arbitrary relations representing abstract (non-perceptual) knowledge.
As an example, the phrase ``the cup on the tray is mine and the fruit on the right is fresh"
conveys notions of \emph{possession} and \emph{being fresh} that are factual. 
The space of facts grows through linguistic interactions where any
previously-unknown property of an object is considered a fact.
Facts can be true of multiple entities simultaneously.
For example, in the statement, ``the fruit and the box on the table are my
snack", the notion of \emph{snack} includes the two indicated objects in the scene.
%
%

A set of \emph{imperative} groundings $\Imp{\Grounding{}}$ describe the objectives or goals that can be
provided to a robot motion planner to create a set of robot motions.
Imperative groundings are characterized by a type of motion (e.g., picking, placing or pointing) 
and a set of spatial constraints (e.g., proximity, intersection or contact) that must be satisfied by the 
executed action. 
%
Further, we include aggregative constraints conveyed as
conjunctive references (e.g., ``the block and the can") or associations (e.g., ``a set of blocks").
Finally, the set of imperative groundings also includes a symbol that conveys the assertion of 
an inferred factual grounding\footnote{As we discuss later in Section \ref{sec:pgm}, an imperative 
grounding associated with an asserted fact is used to propagate factual knowledge.}.
The space of grounding symbols can be cumulatively represented as $\Grounding{}  = 
\Imp{\Grounding{}} \cup \Per{\Grounding{}} \cup \Fact{\Grounding{}}$.
Table \ref{tab:gnd-symbols} lists the representative set of grounding predicates used in this work.
\subsection{Language Grounding with Context}
The problem of understanding or \emph{grounding} a natural language utterance involves relating 
the input language with semantic concepts expressed in the workspace. 
This process involves estimating the probable set of groundings that best convey the intended meaning 
of the perceived language utterance. 
For example, the interpretation for the utterance, ``Pick up the block on the table" can be represented by 
the grounding set \textsc{Block}($o_{1}$) $\wedge$ \textsc{EndEffector}($o_{2}$) $\wedge$ \textsc{Table}($o_{3}$) $\wedge $ \textsc{On}($o_{1}$,$o_{3}$) $\wedge$ \textsc{Contact}($o_{2}$,$o_{1}$) where object symbols $o_{1}$, $o_{2}$ and $o_{3}$ are derived from the symbolic world model. 
%
The estimated association between the input language utterance and the set of grounding symbols 
expresses the intended meaning of the sentence and can be considered as determining a \emph{conceptual} grounding 
for the utterance. 
%
Further, the set of object symbols that parameterize the estimated grounding symbols must be associated with the set of visual percepts arising 
with the geometric objects populating the workspace. This process can be viewed as estimating \emph{existential} groundings.  
In the above example, this would entail associating object symbols $o_{1}$ and $o_{2}$ with 
the set of detections $\VisualObservations{0:t}$ that correspond to the physical block and table. 
In essence, the existential grounding process 
accounts for uncertainty in the robot's perception of semantic entities present in the environment.   

Following \cite{tellex2011understanding}, the grounding process is mediated by a binary correspondence 
variable $\CV{ij} \in \Correspondences{}$ 
that expresses the degree to which the phrase $\Phrase{i} \in \Language{}$ corresponds to a possible 
grounding $\GV{j} \in \Grounding{}$. 
This allows groundings to be expressed as probabilistic predicates
modeling the uncertainty in the degree of association of a concept with a phrase. 
Further, limiting the domain of correspondence variables to a \textit{true} or a \textit{false} association, 
allows the factors relating phrases and candidate groundings to be locally normalized, which reduces the learning complexity. 
Similarly, we introduce correspondence variables 
$\Phi^{\mathcal{O}}$ that convey probable associations between object symbols in the world model $\Upsilon$ 
and visual detections $\VisualObservations{0:t}$. 
%

%
A note on convention: in the remainder of this paper, we use the phrase ``grounding natural language" to imply  
the \emph{conceptual} grounding of an input language utterance to the space of grounding symbols 
that characterize workspace knowledge acquired by the robot. 
We address the issue of estimating correspondences between object symbols in the world model with the visual percepts 
in Section~\ref{subsec:dec-gnd}. 
Further, in our formulation the grounding for an utterance is obtained by determining the likely true correspondences. 
For brevity at times we refer to ``estimating  true correspondences between phrases in the utterance and grounding symbols" as simply ``estimating groundings for language". 
Finally, we use capital symbols in equations to refer to sets of variables.

%
%
%
We now discuss the temporally extended scenario where estimating the grounding for an input language 
utterance involves reasoning over an accrued context of past visual observations and language utterances.
For example, consider the scenario where a robot observes a human place a can in the scene.   
This is followed by the human uttering ``the can that I put down is my snack"
and commanding the robot to ``pack up my snack''. 
Interpreting this sequence of language utterances requires inferring and reasoning with factual information, such as which objects constitute the ``snack'', 
recognizing that the ``put down'' action is associated with the ``can'', and inferring that the robot is instructed to 
execute a ``lift and place'' action sequence to satisfy the command. 
Given the context of visual observations  $\VisualObservations{0:t}$ and utterances $\Language{0:t-1}$ leading up to time $t$, 
the problem of estimating the set of true correspondences $\Correspondences{t}$ for the utterance $\Language{t}$ and control actions $\Control$ can be posed as:
\begin{equation} \label{eq:joint-dist-all-context}
 P( \Control, \Correspondences{t} | \Language{0:t}, \VisualObservations{0:t}, \Grounding{t}). 
\end{equation}
In the following section, we develop a probabilistic model for estimating this distribution.

\section{Probabilistic Model}
\label{sec:pgm}

\begin{figure*}[t]
  \centering
  \scalebox{0.6}{\begin{tikzpicture}
      \node[] (s1) {\large The box I put down on the table is mine.};

      \node[below=of s1, yshift=0.2cm, xshift=-1.5cm] (d1) {\large $\text{Human}({\red x})\wedge\text{PutDown}({\red x},{\green y})\wedge\text{Box}({\green y})\wedge\text{On}({\green y},{\orange z})\wedge\text{Table}({\orange z})$};
      \node[right=of d1, xshift=0.0cm] (i1) {\large $\text{Assert}(\text{{\blue Mine}}({\green y}))$};

      \draw[->,thick,bend left=-15] ($(s1.south)+(-0.1,0)$) to node[left,xshift=-0.4cm]{$\Dec{ \Language{t-1} }$} ($(d1.north)+(0,0)$);
      \draw[->,thick] ($(s1.south)+(2.5,0)$) to[out=-30,in=140] node[right,xshift=0.2cm,yshift=0.2cm]{$\Imp{ \Language{t-1} }$} ($(i1.north)+(0,0)$);

      \node[yshift=-1cm,xshift=-4cm,anchor=south west,inner sep=0,below=of s1] (f0) {\includegraphics[width=0.13\textwidth]{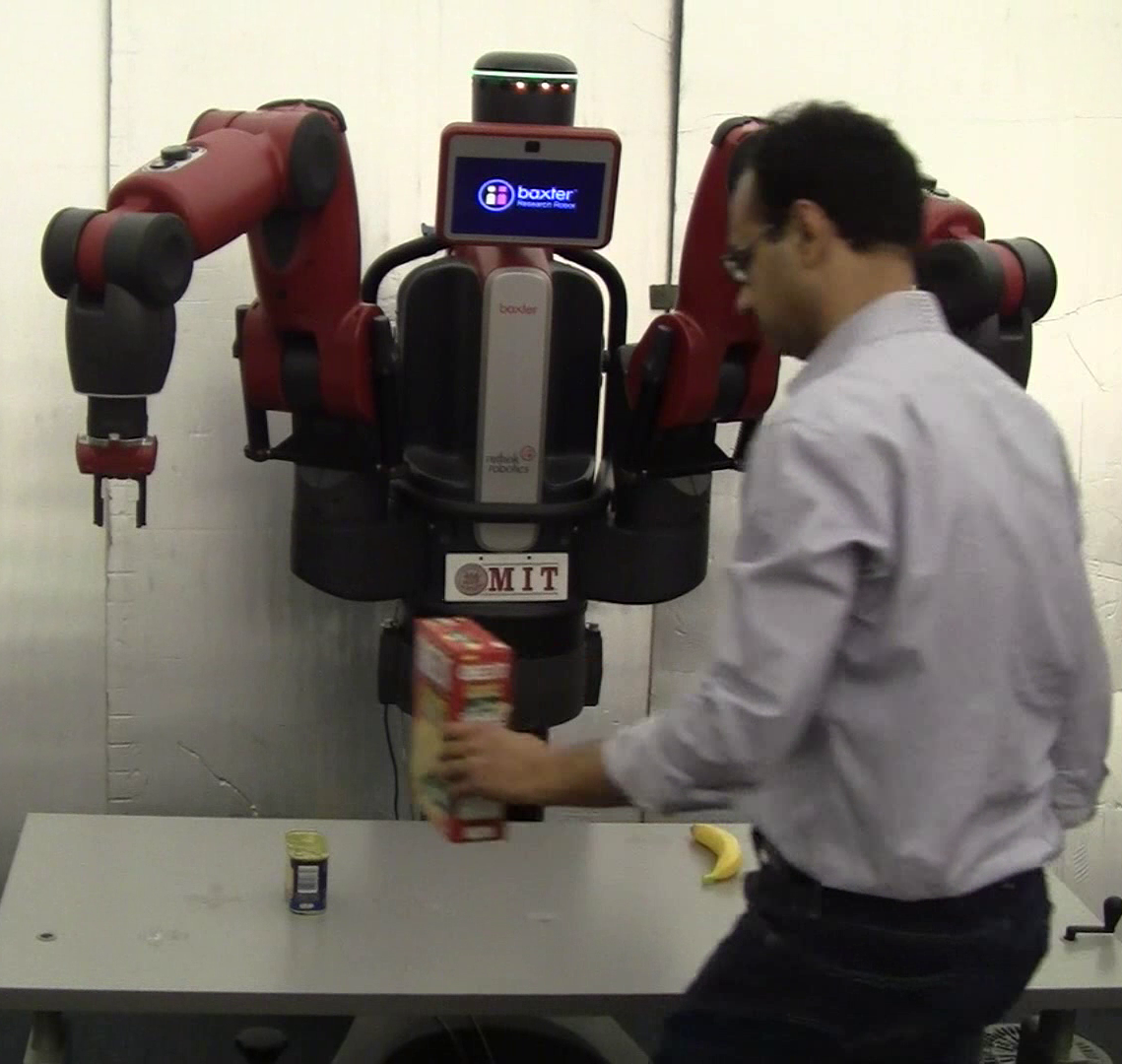}};
      \node[anchor=south west,inner sep=0,right=of f0] (f1) {\includegraphics[width=0.13\textwidth]{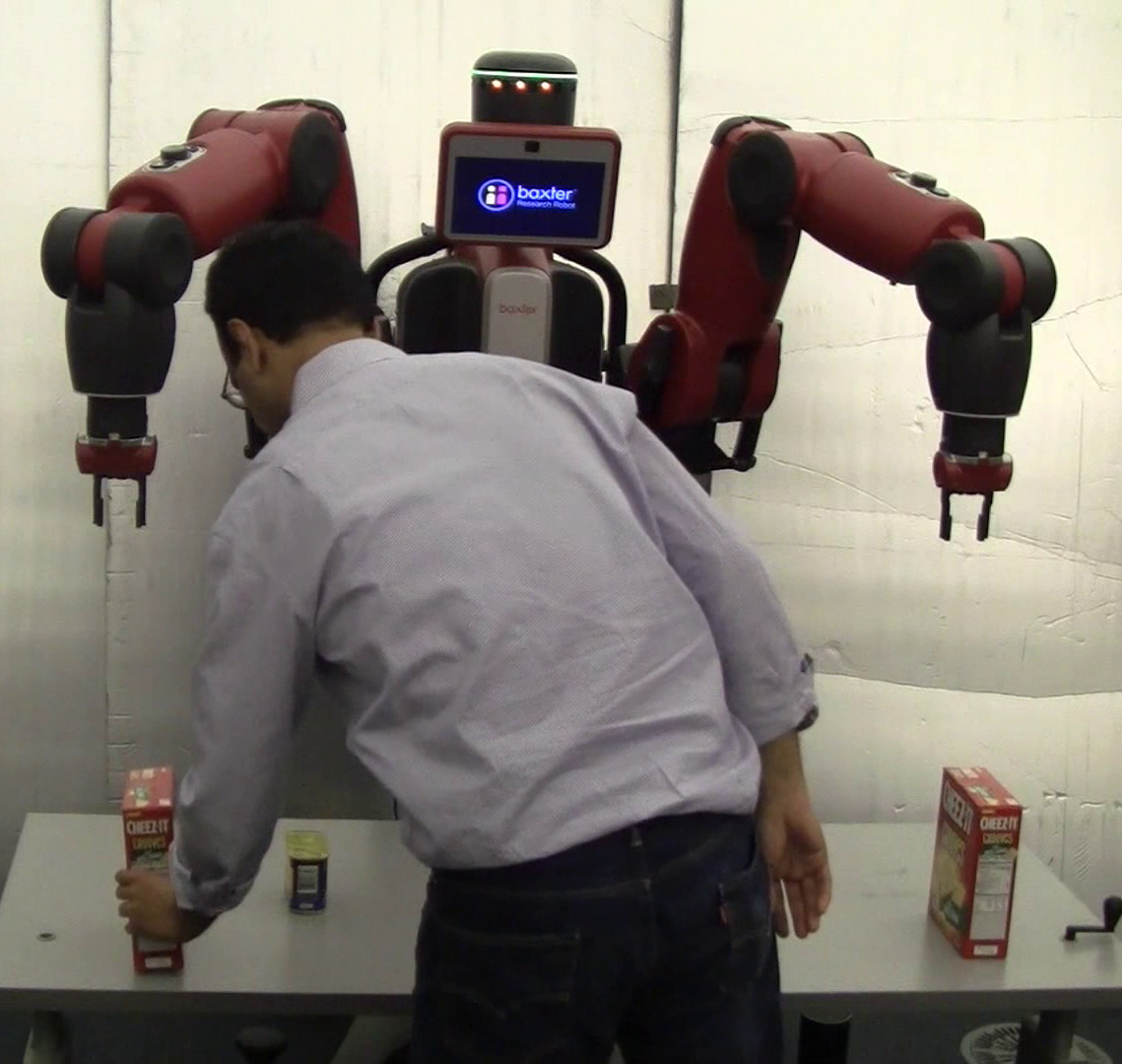}};
      \node[anchor=south west,inner sep=0,right=of f1] (f2) {\includegraphics[width=0.13\textwidth]{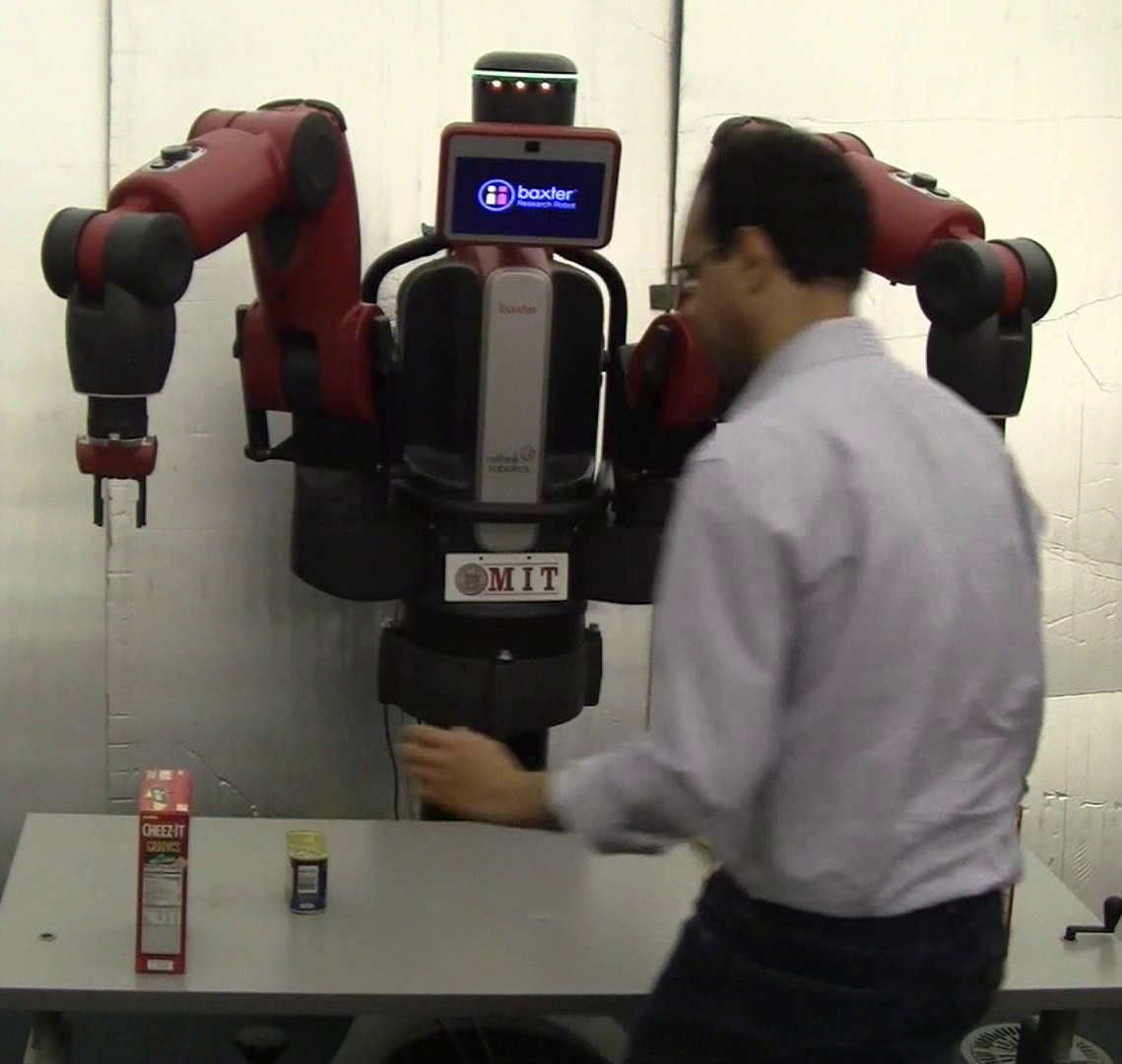}};
      
      \node[yshift=1cm,below=of f1] (lf1) {Visual observations $\VisualObservations{0:t}$};
      \node (p01) at ($(f0)!0.5!(f1)$) {$\cdots$};
      \node (p10) at ($(f1)!0.5!(f2)$) {$\cdots$};
      
      \draw[-,thick,red,dashed] ($(f0.south west)+(2.0,1)$) -- ($(f0.south west)+(4.6,0.85)$) -- ($(f0.south west)+(8.6,1.15)$);
      \draw[-,thick,green,dashed] ($(f0.south west)+(1.0,0.7)$) -- ($(f0.south west)+(3.6,0.5)$) -- ($(f0.south west)+(7.0,0.4)$);
      \draw[-,thick,orange,dashed] ($(f0.south west)+(1.2,0.2)$) -- ($(f0.south west)+(4.6,0.2)$) -- ($(f0.south west)+(8.0,0.2)$);

      \fill[red,opacity=0.1] ($(f0.south west)+(2.0,1)$) -- ($(f0.south west)+(4.6,0.85)$) -- ($(f0.south west)+(8.6,1.15)$) -- ($(d1.south)+(-0.8,0.1)$);
      \fill[green,opacity=0.1] ($(f0.south west)+(1.0,0.7)$) -- ($(f0.south west)+(3.6,0.5)$) -- ($(f0.south west)+(7.0,0.4)$) -- ($(d1.south)+(0.97,0.1)$);
      \fill[orange,opacity=0.1] ($(f0.south west)+(1.2,0.2)$) -- ($(f0.south west)+(4.6,0.2)$) -- ($(f0.south west)+(8.0,0.2)$) -- ($(d1.south)+(2.65,0.1)$);

      \node[right=of i1, yshift=-1.5cm, xshift=-0.9cm] (g) {\includegraphics[width=0.05\textwidth]{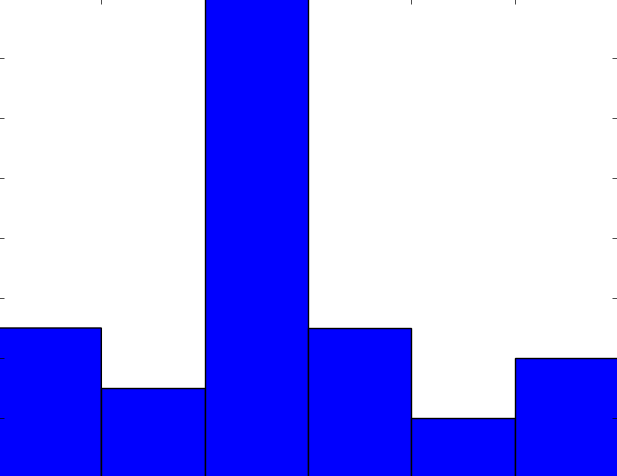}};
      \node[below=of g, yshift=1cm] (gcap) {\large State $\KB{t-1}$};
      
      \draw[->,thick] ($(i1.south)+(0.4,0)$) to[out=-80,in=150] node[left,yshift=-0.4cm,xshift=0.3cm]{update} ($(g.west)+(0,0)$);

      \node[anchor=south west,inner sep=0, right=of f2, xshift=13cm,yshift=1.2cm] (f3)
      {\includegraphics[width=0.28\textwidth]{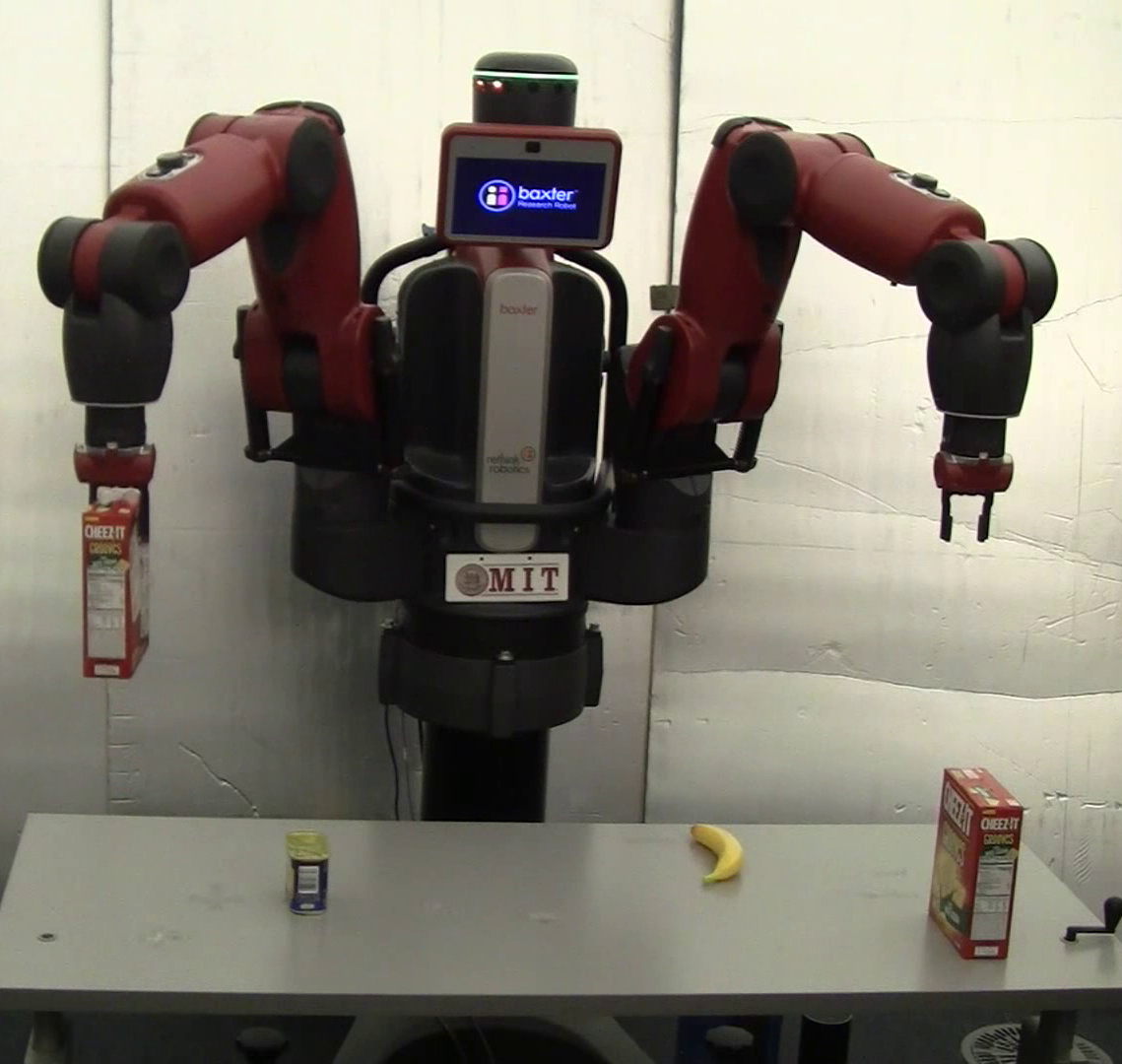}};
      \node[yshift=1cm,below=of f3] (lf3) {Robot performing the intended action};

      \node[right=of s1, xshift=6cm] (s2) {\large Pack up my box.};

      \node[below=of s2, yshift=0.2cm, xshift=-2cm] (d2) {\large ${\blue\text{Mine}}({\violet x})\wedge\text{Box}({\violet x})$};
      \draw[->,thick] ($(g.east)+(0,0)$) to[out=20,in=-100] node[right,yshift=-0.4cm,xshift=-0.2cm]{lookup} ($(d2.south)+(-0.7,0)$);

      \draw[->,thick,bend left=-5] ($(s2.south)+(-0.2,0)$) to node[left,xshift=-0.7cm]{$\Dec{ \Language{t} }$} ($(d2.north)+(0,0)$);

      \node[obs, below=of s2, xshift=1cm, yshift=-2cm] (n0) {$\lambda_1$\strut};
      \node[obs, right=2 of n0] (n1) {$\lambda_2$\strut};
      \node[below=of n0, yshift=1cm] (n0s) {\large Pack up\strut};
      \node[below=of n1, yshift=1cm] (n1s) {\large box\strut};
      
      \node (dcg) [rounded corners, dashed, draw=black, fit= (n0) (n1) (n0s) (n1s), inner sep=0.18cm] {};
      \draw[->,black,thick,shorten >=3pt] ($(s2.south)+(0,0)$) to[out=-90,in=160] node[left,xshift=0cm]{$\Imp{ \Language{t} }$} ($(dcg.west)$);

      \node[obs, above=of n0, yshift=-0cm,minimum size=0.8cm] (g0a) {$\gamma_1$\strut};
      \node[obs, above=of n1, yshift=-0cm,minimum size=0.8cm] (g1a) {$\gamma_2$\strut};
      \node[latent, above=of n0, yshift=-1.2cm,minimum size=0.8cm,xshift=1cm] (p0a) {$\phi_1$};
      \node[latent, above=of n1, yshift=-1.2cm,minimum size=0.8cm,xshift=1cm] (p1a) {$\phi_2$};
      \factor[above=0.5 of n0] {f0n} {} {} {};
      \factoredge {p0a,n0,g0a,g1a} {f0n} {};
      \factor[above=0.5 of n1] {f1n} {} {} {};
      \factoredge {p1a,n1,g1a} {f1n} {};
      
      \draw[-,violet,thin,dashed,shorten >=3pt] ($(d2.north)+(1.3,0)$) to[out=30,in=160] ($(f1n)$);

      \draw[->,thick,shorten >=5pt,shorten <=5pt] ($(p1a.east)+(0,0.3)$) to[out=10,in=220] ($(f3.west)$);

      \fill[violet!60,opacity=0.5,rounded corners, dashed] ($(f3.south west)+(0.3,1.5)$) rectangle ($(f3.south west)+(0.7,3)$);
      \draw[violet,rounded corners, dashed] ($(f3.south west)+(0.3,1.5)$) rectangle ($(f3.south west)+(0.7,3)$);
    \end{tikzpicture}}%
\caption{An overview of grounding commands from the initial utterances to a
  robotic action. Utterances are jointly grounded into declarative and
  imperative grounding symbols. Perceptual groundings are computed jointly with
  object tracks, shown with corresponding colors. Factual
  knowledge is aggregated and used to disambiguate later commands. A
  grounding graph is illustrated for a command showing how the structure of the model mirrors
  that of the sentence. Probable imperative correspondences are
  estimated conditioned on the declarative groundings and prior state. They
  determine the constraints used by a motion planner to generate an action
  sequence and direct the robot's behaviour.}%
  \label{fig:overview}
\end{figure*}
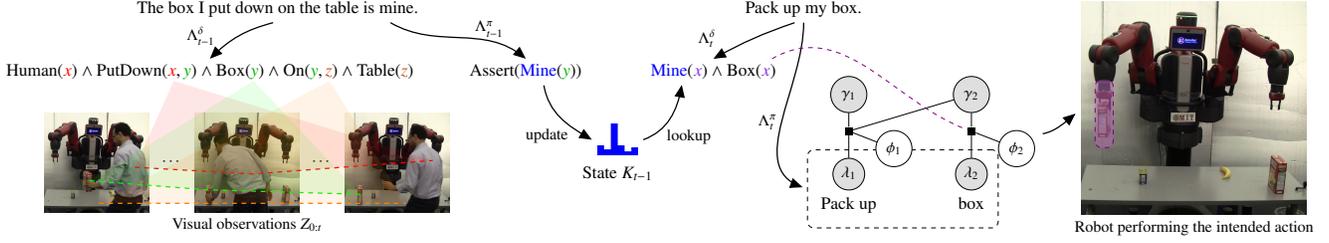

In this section, we present the \emph{Temporal Grounding Graphs} model for interpreting natural language utterances with past visual-linguistic context of observations of the world coupled 
with descriptive language. 
We pose the problem of estimating Equation \ref{eq:joint-dist-all-context} as ``filtering" on a 
dynamic Bayesian network and subsequently detail the model for estimating groundings for an utterance at each time step.
Further, we discuss model training and present an analysis for runtime and space complexity for the model. 
\newcommand{\subdiag}[7]{%
  \node[#7, #1=of #2t, xshift=#3cm, yshift=#4cm] (#5t) {$#6{t}$};
}
\newcommand{\subdiagX}[7]{%
  \node[#7, #1=of #2t, xshift=#3cm, yshift=#4cm] (#5t) {$#6{0:t}$};
}


\subsection{Temporal Model} 
\renewcommand{\subdiag}[7]{%
  \node[#7, #1=of #2t1, xshift=#3cm, yshift=#4cm] (#5t1) {$#6{1}$};
  \node[#7, #1=of #2tn1, xshift=#3cm, yshift=#4cm] (#5tn1) {$#6{t-1}$};
  \node[#7, #1=of #2t, xshift=#3cm, yshift=#4cm] (#5t) {$#6{t}$};
}
\tikzset{
    between/.style args={#1 and #2}{
         at = ($(#1)!0.5!(#2)$)
    }
  }

\begin{figure}[t]
\center
  \centering
  \scalebox{0.48}{\begin{tikzpicture}
    \node[obs] (kb0) {$\KB{0}$};
    \node[latent, right=of kb0, xshift=1.6cm] (kb1) {$\KB{1}$};
    \node[latent, right=of kb1, xshift=1.6cm] (kbtn2) {$\KB{t-2}$};
    \node[latent, right=of kbtn2, xshift=1.6cm] (kbtn1) {$\KB{t-1}$};
    \node[latent, right=of kbtn1, xshift=1.6cm] (kbt) {$\KB{t}$};
    \factor[between=kb0 and kb1, xshift=0.1cm] {ft1} {} {} {};
    \factor[between=kbtn2 and kbtn1, xshift=0.1cm] {ftn1} {} {} {};
    \factor[between=kbtn1 and kbt, xshift=0.1cm] {ft} {} {} {};
    \draw[dashed] (kb1) -- (kbtn2);
    \subdiag{below}{f}{0}{0}{c}{\Correspondences}{latent}
    \factoredge {kb0, kb1, ct1} {ft1} {};
    \factoredge {kbtn2, kbtn1, ctn1} {ftn1} {};
    \factoredge {kbtn1, kbt, ct} {ft} {};
    \subdiag{below}{f}{-0.8}{0.6}{gr}{\Grounding}{obs}
    \subdiag{below}{c}{1.2}{0.6}{ac}{\Action}{latent}
    \subdiag{below}{c}{-0.2}{0.3}{la}{\Language}{obs}
    \subdiag{below}{c}{-2.0}{-0.8}{vo}{\VisualObservations}{obs}
    \factor[below=0.8 of grt1, xshift=0.0cm] {mt1} {} {} {};
    \factor[below=0.8 of grt, xshift=0.0cm] {mt} {} {} {};
    \factor[below=0.8 of grtn1, xshift=0.0cm] {mtn1} {} {} {};
    \factoredge {grt1,lat1,vot1,ct1,kb0} {mt1} {};
    \factoredge {grt,lat,vot,ct,kbtn1} {mt} {};
    \factoredge {grtn1,latn1,votn1,ctn1,kbtn2} {mtn1} {};
    \factor[right=0.1 of ct1, yshift=-0.5cm] {pl1} {} {} {};
    \factor[right=0.1 of ct, yshift=-0.5cm] {pl} {} {} {};
    \factor[right=0.1 of ctn1, yshift=-0.5cm] {pln1} {} {} {};
    \factoredge {ct1,act1} {pl1} {};
    \factoredge {ct,act} {pl} {};
    \factoredge {ctn1,actn1} {pln1} {};
    \node (gt1) [rounded corners, dashed, draw=red, fit= (vot1) (votn1), inner sep=0.12cm] {};
    \node (gt) [rounded corners, dashed, draw=red, fit= (vot1) (vot) (votn1), inner sep=0.18cm] {};
  \end{tikzpicture}}
  \caption{
  \footnotesize{
  Grounding an instruction $\Language{t}$ with past visual-linguistic observations $ \{ \VisualObservations{0:t}, 
  \Language{0:t-1} \}$ is posed as filtering on a temporal model. A state variable $\KB{t}$ provides 
  minimal state persistence over factual knowledge derived from past language inputs $\Language{0:t-1}$. 
  Estimating perceptual groundings from past visual observations $\VisualObservations{0:t}$ is delayed, variables 
  encircled with red plate. 
  The declarative context within the instruction $\Language{t}$ focuses inference over past visual observations.}
  }
  \label{fig:temporal-model}
\end{figure}
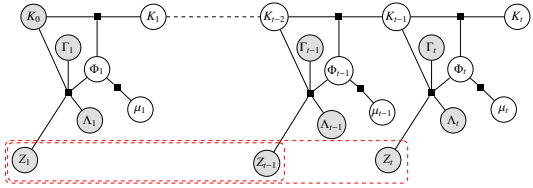
We pose the problem of computing Equation \ref{eq:joint-dist-all-context} 
as incremental estimation on a temporal model.
The interpretation of a current utterance $\Language{t}$ can rely on the  
background knowledge derived from the past language descriptions $\Language{0:t-1}$ or 
semantic information derived from accrued visual observations $\VisualObservations{0:t}$.
As a result, conditional dependencies exist between the interpretation for current utterance and past observations.
The direct estimation of this distribution is challenging since the  
interpretation of the current command $\Language{t}$ may involve reference to past knowledge 
that must be estimated jointly from past observations, creating conditional dependencies that grow with time.
One approach is to introduce a state variable incorporating inferred past groundings that is propagated over time. 
An explicit state variable has the advantage of making the past observations conditionally independent 
of future observations given the current state, but
such a variable must maintain all perceptual groundings derived from the history of the visual input. 
The space of perceptual groundings $\Per{\Gamma_{t}}$ is exponentially large
as it must include all possible relationships between entities in the scene that may be referenced by 
a language utterance and is infeasible to maintain for complex scenes over time. 
Alternatively, the model can accrue observations and directly estimate Equation \ref{eq:joint-dist-all-context} 
at each time step. 
Although this approach reduces the space complexity for maintaining context, the 
cost of incremental inference is high since the current and past language utterances must be interpreted jointly. 

We seek to balance the complexity of context maintenance with efficiency of incremental estimation 
and introduce a state variable denoted by $\KB{t}$ that expresses the belief over factual grounding symbols $\Fact{\Gamma_{t}}$.
We assume that factual groundings are uncorrelated. The likelihood over the state can be factored as a product of individual grounding likelihoods as:
%
%
\begin{equation} \label{eq:state-facts}
\KB{t} = 
\bigg\{ 
  \gamma^{ \GroundingType}_{ \Time{\Role}} 
  \hspace{1mm} | \hspace{1mm} 
  \forall \GroundingType \in \Fact{\Predicates}, 
  \forall \Time{\Role} \subseteq \Time{\SymWM}
\bigg\} 
 \hspace{1mm}
P\left( \KB{t} \right) = 
  \prod_{i = 1}^{ | \Fact{\Grounding{t}} |}
  P\left( \gamma^{ \GroundingType}_{\Time{\Role}i} \right).
\end{equation}
Note that the likelihood of the 
state variable expresses the likelihood \emph{over} the factual grounding variables. 
This is in contrast to the grounding likelihood that explicitly models the degree of association \emph{between} an utterance and a factual concept. 
The inclusion of the state variable in Equation \ref{eq:joint-dist-all-context} at each time step 
enables the following factorization:
\begin{equation} \label{eq:past-present-separation}
\begin{split}
 P( \Control, \Correspondences{t}, \KB{t} | \Language{0:t}, \VisualObservations{0:t}, \Grounding{t}) ={}&  \\
 \Sum{ \KB{t\text{-}1} }
 \overbrace{
 P( \Control,\Correspondences{t}, \KB{t} | \Language{t}, \VisualObservations{0:t}, \KB{t\text{-}1}, \Grounding{t} )
 }^{\text{Current inference}} &
 \overbrace{
 P( \KB{t\text{-}1} | \Language{0:t\text{-}1}, \VisualObservations{0:t\text{-}1}, \Grounding{t})
 }^{\text{Previous state estimate}}.
\end{split}
\end{equation}
The state variable $\KB{t-1}$ represents the cumulative belief over factual groundings till the 
previous time step $t-1$. The state variable $\KB{t-1}$ decouples the estimation of the 
grounding $\Correspondences{t}$ for the current command $\Language{t}$ from the past language 
inputs $\Language{0:t-1}$. 
The factorization introduced in Equation \ref{eq:past-present-separation} poses the inference over the full  
time history as filtering at each time step over groundings for the command and associated actions. 
Further, the online estimation process also propagates or updates the state variable informed by the visual and 
language observations. Figure \ref{fig:temporal-model} illustrates the unrolled temporal model. 
%
%
%
Crucially, we only propagate factual groundings $\Fact{\Grounding{t}}$. %
The set of visual detections $\VisualObservations{0:t}$ are stored directly as part of context. 
The inference over perceptual groundings $\Per{\Grounding{t}}$  
is delayed till a future utterance is received, a process we outline next.

\subsection{Grounding Network}
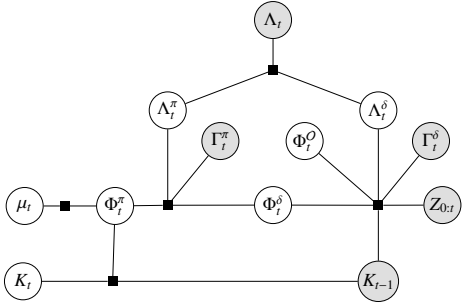
\begin{figure}[]
  \centering
  \scalebox{0.7}{\begin{tikzpicture}
    \node[obs] (lang) {$\Language{t}$};
    \node[latent, below=of lang, xshift=-2cm] (imp) {$\Imp{ \Language{t} }$};
    \node[latent, below=of lang, xshift=2cm] (dec) {$\Dec{ \Language{t} }$};
    \factor[below=of lang, yshift=-0.1cm] {start} {} {} {};
    \factoredge {imp, dec, lang} {start} {};
    %
    \node[latent, below=2.8 of lang] (deccor) {$\Dec{ \Correspondences{t} }$};
    \node[latent, above=0.5 of deccor, xshift=0.6cm] (objcor) { $\Correspondences{t}^{\mathcal{O}}$ } ;
    \node[latent, below=1.1 of imp, xshift=-1cm] (impcor) {$\Imp{ \Correspondences{t} }$};
    \node[obs, right=2.5 of deccor] (vision) {$\VisualObservations{0:t}$};
    \node[obs, above=0.5 of deccor, xshift=3cm] (worldD) {$\Dec{\Grounding{t}}$};
    \node[obs, above=0.5 of deccor, xshift=-1cm] (worldI) {$\Imp{\Grounding{t}}$};
    \node[obs, below=2.5 of dec] (kb) {$\KB{t-1}$};
    \factor[below=1.35 of dec] {tracker} {} {} {};
    \factoredge {vision, worldD, kb, deccor, objcor, dec} {tracker} {};
    \factor[below=1.35 of imp] {dcg} {} {} {};
    \factoredge {imp, deccor, worldI, impcor} {dcg} {};
    \node[latent, left=6 of kb] (kbnext) {$\KB{t}$};
    \factor[right=1.25 of kbnext] {kbupdate} {} {} {};
    \factoredge {kb, impcor, kbnext} {kbupdate} {};
    \node[latent, left=1 of impcor] (action) {$\Action{t}$};
    \factor[left=0.5 of impcor] {planner} {} {} {};
    \factoredge {action, impcor} {planner} {};
  \end{tikzpicture}}
\caption{ 
 \footnotesize{
 The grounding network for inference at each time instant $t$.
 An utterance $\Language{t}$ is grounded through imperative correspondences, $\Imp{\Correspondences{t}}$,
 related to future robot actions, and declarative correspondences, $\Dec{\Correspondences{t}}$, relating perception and factual knowledge. 
 A parsing mechanism facilitates this by partitioning $\Language{t}$ into $\{ \Imp{\Language{t}}, \Imp{\Language{t}} \}$. 
 The declarative grounding factor incorporates visual observations $\VisualObservations{0:t}$ and prior state estimate $\KB{t-1}$.  
 Estimation of imperative groundings is conditioned on declarative correspondences and yields constraints for planning robot actions $\Control$. 
 Estimated groundings update and propagate the state estimate $\KB{t}$.}
 }
 \label{fig:ground-network}
\end{figure}
We now detail the model for inference instantiated at each time step within the dynamic Bayesian network outlined in the previous section.
The grounding model estimates the likely set of correspondences $\Correspondences{t}$, the control sequence $\Control$ and the 
updated knowledge state $\KB{t}$ from the utterance $\Language{t}$ at time $t$ using the prior state $\KB{t-1}$ and visual observations $\VisualObservations{0:t}$.
We assume that the estimated groundings of an input command are sufficient for determining the output robot actions. 
The inference over the control sequence $\Control$ is decoupled from other variables given the knowledge of expressed correspondences $\Correspondences{t}$.
%
%
Further, we assume that given the prior state $\KB{t-1}$ and the observed correspondences $\Correspondences{t}$, the updated state variable $\KB{t}$ is conditionally independent of other variables in the model. 
The above assumptions factorize the joint likelihood as: 
\begin{equation} \label{eq:current-model-structure}
\begin{split}
 P( \Control, \Correspondences{t}, \KB{t} | \Language{t}, \VisualObservations{0:t}, \KB{t\text{-}1}, \Grounding{t} ) =&{}  \\
 \overbrace{ P( \Control | \Correspondences{t}) }^{ \text{Planner} } 
 \overbrace{ P( \KB{t} | \Correspondences{t}, \KB{t\text{-}1} ) }^{ \text{State update} } 
& \overbrace{ P( \Correspondences{t} | \Language{t}, \VisualObservations{0:t}, \KB{t-1}, \Grounding{t}) }^{ \text{Grounding model} }.
\end{split} 
\end{equation}
In the above, the grounding model  $P( \Correspondences{t} | \Language{t}, \VisualObservations{0:t}, \KB{t-1}, \Grounding{t})$ estimates the 
correspondences for the input instruction $\Language{t}$ given context accrued till the present time step $\{ \KB{t-1}, \VisualObservations{0:t} \}$.
Note that an input language utterance may instruct the robot to perform a task, refer to past or current knowledge about the workspace or provide 
factual knowledge for future inferences. 
Estimating the conditional likelihood $P( \Correspondences{t} | \Language{t}, \VisualObservations{0:t}, \KB{t-1}, \Grounding{t})$ 
involves determining the set of true imperative correspondences $\Imp{ \Correspondences{t} }$, the declarative correspondences $\Dec{ \Correspondences{t} }$ 
and the set of correspondences $\Correspondences{t}^{\mathcal{O}}$ that object symbols in the world model $\SymWM$ with visual percepts $\VisualObservations{0:t}$.  
Partitioning the set of grounding variables as imperative and declarative sets allows Equation \ref{eq:current-model-structure} as:   
\begin{equation} \label{eq:imp-dec}
\begin{split}
 P( \Control, \Correspondences{t}, \KB{t} | \Language{t}, \VisualObservations{0:t}, \KB{t-1}, \Grounding{t}) =   
 \Sum{ \Imp{\Language{t}}, \Dec{\Language{t}} }
 \overbrace{
 P( \Control | \Imp{ \Correspondences{t} } )
 }^{\text{Planner}}
 \overbrace{
 P( \KB{t} | \Imp{ \Correspondences{t} }, \KB{t\text{-}1} )
 }^{\text{State update}} \\
 \underbrace{ 
 P( \Imp{ \Correspondences{t} } | \Imp{ \Language{t} } , \Dec{ \Correspondences{t} }, \Imp{\Grounding{t}} )
 }_{ \text{Imperative grounding }}
 \underbrace{
 P( \Dec{ \Correspondences{t} }, \Correspondences{t}^{\mathcal{O}} | \Dec{ \Language{t} }, \VisualObservations{0:t}, \KB{t\text{-}1}, \Dec{\Grounding{t}})
 }_{ \text{Declarative grounding } }
 \underbrace{
 P( \Imp{ \Language{t}}, \Dec{ \Language{t}} | \Language{t} )
 }_{ \text{Parsing } }.
\end{split}
\end{equation}
Figure \ref{fig:ground-network} illustrates the resulting factor graph.  
The declarative grounding factor $P( \Dec{ \Correspondences{t} }, \Correspondences{t}^{\mathcal{O}} | \Dec{ \Language{t} }, \VisualObservations{0:t}, \KB{t\text{-}1}, \Dec{\Grounding{t}})$ 
involves estimating perceptual and factual groundings conditioned on the context of visual observations and the propagated knowledge state. 
Further, this factor jointly estimates the likely correspondences that connect the symbols in the world model with percepts obtained by the visual sensor.
%
The estimation of imperative groundings is conditioned on the declarative grounding variables described by the 
likelihood $P( \Imp{ \Correspondences{t} } | \Imp{ \Language{t} } , \Dec{ \Correspondences{t} }, \Imp{\Grounding{t}})$. 
The declarative grounding variables decouple the estimation of the imperative groundings from the 
past visual observations and the current knowledge state.
The inclusion of declarative grounding variables in the factor estimating the imperative grounding enables disambiguation of the 
future actions that the robot may perform. 
For example, consider the utterance, ``lift the box that was placed by the person on the table" where the intended manipulation of the box
is constrained by the knowledge of a past interaction between the person and the specific box.
The induced factorization in Equation \ref{eq:imp-dec} relies on partitioning a set of linguistic constituents as
imperative (future actions) or declarative (knowledge from the past). Parsing of the instruction and estimating a partitioning of the constituents set as 
imperative or declarative is modeled by the factor $P(\Imp{ \Language{t}}, \Dec{ \Language{t}} | \Language{t})$.

We assume that the estimated imperative groundings derived from a language command fully characterize the goals and objectives for future 
robot actions. The estimation of the control sequence generation is conditioned on the estimated imperative groundings alone 
expressed in the factor $P(\Control|\Imp{\Correspondences{t}})$. 
If the language utterance conveys factual information about entities in the world, then the estimated factual groundings for the utterance serve as a ``measurement" for knowledge state.
%
%
The estimated imperative grounding $\Imp{\Correspondences{t}}$ involves an action to update the prior knowledge $\KB{t-1}$ using the 
estimated factual groundings to form the propagated state $\KB{t}$, modeled in the likelihood $P( \KB{t} | \Imp{ \Correspondences{t} }, \KB{t\text{-}1})$.

\subsection{Factors}
\label{subsec:factors}
In this section, we detail the individual models that constitute the ground network represented by Equation \ref{eq:imp-dec}. 
%
\subsubsection{Dependency Parse and Syntactic Analysis}
The parsing model $P( \Imp{ \Language{t} }, \Dec{ \Language{t} } | \Language{t} )$ is realized 
using a dependency parser, START \cite{katz1988using}, 
that factors the input language $\Language{t}$ into a set of constituents related to each other using syntactic dependency relationships.
The structure of the computed linguistic relations informs the structure of the grounding models 
such that conditional dependencies between grounding variables follow the parse structure.
The set of constituents and syntactic dependencies, as well as the morphological
and lexical features provided by START, determine if a constituent plays an
imperative $\Imp{\Language{t}}$ and/or declarative $\Dec{\Language{t}}$ role in
the input utterance.
This partitioning informs the grounding factors $P( \Imp{ \Correspondences{t} } | \Imp{ \Language{t} } , \Dec{ \Correspondences{t} }, \Imp{\Grounding{t}})$ 
and $P( \Dec{ \Correspondences{t} }, \Correspondences{t}^{\mathcal{O}}  | \Dec{ \Language{t} }, \VisualObservations{0:t}, \KB{t\text{-}1}, \Dec{\Grounding{t}})$. 
Any linguistic constituents which are unknown are ignored aside from those associated with a single entity.
These are assumed to be new facts about that entity which enables the corpus of
factual knowledge to expand online.

\subsubsection{Declarative Grounding Model} 
\label{subsec:dec-gnd}
The declarative grounding model estimates the set of correspondences for declarative groundings $\Dec{\Correspondences{t}}$.
These include  
facts, static spatial relations, and dynamic relations such as events.
Further, the factor also estimates the correspondences $\Correspondences{t}^{\mathcal{O}}$ 
between objects symbols and percepts. 
The estimation is conditioned on the set of declarative linguistic constituents
$\Dec{\Language{t}}$ derived from the parse of an utterance, the set of visual
observations $\VisualObservations{0:t}$ and the prior state $\KB{t-1}$.
Since factual relations express abstract or non-perceptual knowledge, 
the set of factual groundings is decoupled from visual observations conditioned 
on the utterance and the prior state. 
This enables the following factorization where we assume an undirected representation: 
\begin{equation} \label{eq:dec-fact-a}
\resizebox{.9\hsize}{!}{\begin{math}
\begin{split}
\Psi( \Dec{ \Correspondences{t} }, \Correspondences{t}^{\mathcal{O}}, \Dec{ \Language{t} } , \VisualObservations{0:t}, \KB{t-1}, \Dec{\Grounding{t}} ) &=& 
    \overbrace{
    \Psi( \Correspondences{t}^{\mathcal{F}}, \Dec{\Language{t}}, \KB{t-1}, \Fact{\Grounding{t}})
     }^{\text{Factual grounding}}
    \overbrace{
    \Psi( \Correspondences{t}^{\mathcal{P}},  \Correspondences{t}^{\mathcal{O}}, \Dec{ \Language{t}}, \VisualObservations{0:t}, \Grounding{t}^{\mathcal{P}} )
     }^{\text{Perceptual grounding}}.
\end{split}
\end{math}}
\end{equation}

The factor $\Psi( \Correspondences{t}^{\mathcal{F}}, \Dec{\Language{t}}, \KB{t-1}, \Fact{\Grounding{t}})$ models the likelihood of a factual grounding.
The conditional likelihood is inferred from language $\Fact{\Language{t}}$ and the state variable $\KB{t-1}$ which includes the estimated belief over factual knowledge propagated from a past context.

The factor $\Psi( \Correspondences{t}^{\mathcal{P}},  \Correspondences{t}^{\mathcal{O}}, \Dec{ \Language{t}}, \VisualObservations{0:t}, \Grounding{t}^{\mathcal{P}} )$ 
models the joint likelihood over perceptual groundings $ \Correspondences{t}^{\mathcal{P}}$ and correspondences between world model symbols and percepts $ \Correspondences{t}^{\mathcal{O}}$ 
given language $\Dec{\Language{t}}$ and accrued visual observations $\VisualObservations{0:t}$.
This factor is realized using a vision-language framework developed by \citet{barbu2013saying}, the \emph{Sentence Tracker}.
It models the correspondences between visual observations and an event described
by a sentence as a factorial hidden Markov model \cite{ghahramani1997factorial}.
Given a parse of an input sentence it first determines the number of
participants in the event described by that sentence.
Next, it instantiates one tracker HMM for each participant.
Tracker HMMs model sequences of object detections that both have high likelihood
and are temporally coherent and observe the visual input
$\VisualObservations{0:t}$ to estimate object tracks from raw detections.
Tracker HMMs are in turn observed by declarative grounding HMMs.
An HMM for a declarative symbol encodes the semantics of that symbol as a model
for one or more sequences of detections; for example \emph{approach} can be
modeled as an HMM with three states: first, the objects are far apart, then they
are closing, and finally they are close together.
Grounding models observe trackers according to the parse structure of the sentence
thereby encoding subtle differences in meaning, like the difference between
being given a block and giving away a block.
For example, given the sentence ``The person put down the block on the table''
and a parse of that sentence as \textsc{Human}$(x)$ $\wedge$ \textsc{Block}$(y)$
$\wedge$ \textsc{PutDown}$(x,y)$ $\wedge$ \textsc{Table}$(z)$ $\wedge$
\textsc{On}$(y,z)$ it instantiates three trackers, one for each variable, and
five declarative grounding models, one for every predicate, and connects them
according to the predicate argument structure of the parse.
Inference simultaneously tracks and recognizes the sentence finding the
globally-optimal object tracks for a given sentence and set of object
detections.
The formulation described above expresses Equation \ref{eq:dec-fact-a} as: 
\begin{equation} \label{eq:dec-fact-b}
\begin{split}
\Psi( \Dec{ \Correspondences{t} }, \Correspondences{t}^{\mathcal{O}}, \Dec{ \Language{t} } , \VisualObservations{0:t}, \KB{t-1}, \Dec{\Grounding{t}} ) = 
    \overbrace{
    \Psi( \Correspondences{t}^{\mathcal{F}}, \Dec{\Language{t}}, \KB{t-1}, \Grounding{t}^{\mathcal{F}})
     }^{\text{Factual grounding}} \\
    \overbrace{
    \Psi( \Correspondences{t}^{\mathcal{P}}, \Dec{ \Language{t}}, \Correspondences{t}^{\mathcal{O}}, \Grounding{t}^{\mathcal{P}} )
     }^{\text{Event grounding}} 
    \overbrace{
    \Psi( \Correspondences{t}^{\mathcal{O}}, \VisualObservations{0:t} )
     }^{\text{Object tracking}}. 
\end{split}
\end{equation}
It is exactly this inference over correspondences $\{ \Dec{\Correspondences{t}}, \Correspondences{t}^{\mathcal{O}} \}$ 
to groundings that allows us to forget the raw observations when grounding future instructions that refer to events in the past. 
\subsubsection{Imperative Grounding Model}
We now discuss the factor $P( \Imp{ \Correspondences{t} } | \Imp{ \Language{t} }, \Dec{ \Correspondences{t} }, \Imp{\Grounding{t}} )$ 
that estimates the correspondences $\Imp{\Correspondences{t}}$ for the imperative instruction given the imperative part of the utterance $\Imp{ \Language{t} }$ 
and the set of declarative correspondence $\Dec{ \Correspondences{t} }$ described previously. 
These sets of imperative groundings include the goals and constraints that are provided to a robot planner to generate the motion plan to satisfy the human's intent.
This factor is realized by extending the \emph{Distributed Correspondence Graph} (DCG) formulation \cite{howard2014efficient,Paul-RSS-16}
which efficiently determines the goal objectives (the object(s) to act upon)
and motion constraints (contact, proximity, visibility etc.) from natural-language instructions.
For example, an utterance like ``lift the farthest block on the right'' results
in contact constraints with one object, a block, which displays the spatial
properties, ``it's on the right", and relations, ``it's the furthest one", implied by
the sentence.
Further, the model allows estimation of aggregate constraints implied in
sentences like ``pick up the can and the box''.

Formally, the imperative grounding likelihood can be expressed in the undirected form as: 
\begin{equation} \label{eq:imp-cv-full}
\begin{split}
\Psi( \Imp{ \Correspondences{t} }, \Imp{ \Language{t} }, \Dec{ \Correspondences{t} }, \Imp{\Grounding{t}} ) = 
  \prod_{i = 1}^{ | \Imp{\Language{t}} | }
  \prod_{j = 1}^{ | \Imp{ \Grounding{} } | }
    \overbrace{
    \Psi( \Imp{ \CV{ij} }, \Imp{ \Phrase{i} }, \Imp{ \GV{ij} }, \Imp{ \Correspondences{c_{ij} } }, \Dec{ \Correspondences{t} })}^{\text{Imperative groundings}}.
\end{split}
\end{equation}
The joint distribution factors hierarchically over the set of linguistic constituents $\lambda_{i} \in \Imp{\Language{t}}$ as 
determined by a syntactic parser. 
The set of linguistic constituents are arranged in a topographical order implied by the 
syntactic relations in the utterance. 
The structure informs the factorization of the joint distribution 
over individual factors where grounding for a linguistic constituent is conditioned on the 
estimated correspondences $\Imp{\Correspondences{c_{ij}}}$ for ``child" constituents that appear earlier in 
the topological ordering. 
Importantly, the conditioning on true declarative correspondences
$\Dec{\Correspondences{t}}$ from earlier constituents couples the estimation of
imperative and declarative groundings for an input instruction.
This probabilistic linkage allows disambiguation of action objectives based on stated declarative knowledge in the 
instruction.
For example, given the utterance ``pick up the block that the human put down'' the observed actions of the human disambiguate which object should be manipulated.

Factors in Equation~\ref{eq:imp-cv-full} are expressed as log-linear models with feature functions that exploit lexical cues, spatial characteristics and the context of child groundings. 
Inference is posed as a search over binary correspondences and is executed by beam search.
The ordered sequence of inferred groundings serve as an input to a planner that generates a robot-specific motion plan $\Control$ to satisfy the inferred objective $\Imp{\Correspondences{t}}$.

\subsubsection{State Propagation}
The state variable $\KB{t}$ maintains a belief over factual groundings; see
Equation~\ref{eq:state-facts}.
It expresses the degree to which a factual attribute is true of an object.
The support for factual grounding variables ranges over workspace entities.
For example, the grounding for an utterance such as ``the block on the table is
mine'' informs the degree to which the fact, the possessive \textsc{Mine}, is
true for the entities which ``block'' is grounded to.
The grounding likelihood can be viewed as an observation from language, informative of the latent 
belief over the factual grounding contained in the knowledge state, and is used to update the propagated 
state.
The factor $P( \KB{t} | \Imp{\Correspondences{t}}, \KB{t-1} )$ models the updated state variable
linked with the grounding obtained for the current utterance with the previous state estimate forming the
prior.
Since factual groundings are assumed to be uncorrelated\footnote{%
  In general one would want knowledge to be structured and to include an
  inference mechanism to deduce consequences and ensure consistency.
  This remains part of future work.}, the posterior distribution over each grounding variable 
in the state is updated independently using a Bayes filter initialized with a uniform prior.
This permits the belief over stored facts to evolve over time providing
resilience to errors and ambiguities as well as accounting for changes in the
environment.

\subsection{Model Training}
\label{subsec:training}
The imperative and declarative factors that constitute the model are trained\footnote{Each factor is trained
  independently using labeled ground truth data. An EM-style approach for
  jointly training all factors remains future work.}
through a data-driven process while the parsing factor used an existing
rule-based model.

The imperative grounding factor, realized using the \emph{Distributed
  Correspondence Graph} (DCG) model, was trained using an aligned corpus of
language instructions paired with scenes where the robot performs a manipulation
task.
A data set consisting of 51 language instructions paired with randomized world configurations
generating a total of 4160 examples of individual constituent-grounding factors.
Ground truth was assigned by hand.
A total of $1860$ features were used for training.
Parameters were trained using a quasi-Newton optimization procedure. 
%
The declarative grounding factor, realized using the \emph{Sentence Tracker},
was trained using captioned videos without any annotation about what the
captions or the words that comprise those actions referred to in the video.
An EM-like algorithm acquired the parameters of the declarative grounding factor
using a corpus of 15 short videos, 4 seconds long, of agents performing actions
in the workspace.
The parsing factor was realized using \emph{START}, a natural language
processing system which is primarily used for question answering from
semi-structured sources such as the World Factbook and Wikipedia.
START was unchanged and we used its standard APIs for language parsing and generation
and for executing actions in response to user requests.

\subsection{Complexity Analysis}
\label{sec:ca}

\begin{table*}[t!]
  \begin{center}
    \vspace{2.2ex}\caption[h]{$O$-bounds on the asymptotic complexity of different approaches to state keeping with Temporal Grounding Graphs.}
    \begin{footnotesize}
    \begin{tabular}{@{}l@{\hspace{5ex}}c@{\hspace{5ex}}c@{\hspace{5ex}}c@{}} 
      & No state keeping & Proposed approached & Full state keeping \\ 
      Observations at each inference step & $\{ \Language{0:t}, \VisualObservations{0:t} \}$ & $\VisualObservations{0:t}$ & $\varnothing$  \\
      State maintained after each inference & $\varnothing$ & $\Gamma^{\mathcal{F}}$ & $\Gamma^{\mathcal{P}}\cup\Gamma^{\mathcal{F}}$ \\
      Space complexity & $\mathbb{\Lambda}t+\mathbb{Z}t$ & $\mathbb{\Gamma}^{\mathcal{F}}\mathbb{o}t$ & $\mathbb{\Gamma}^{\mathcal{P}}\mathbb{o}^{C_{\Lambda}C_{w}}t+\mathbb{\Gamma}^{\mathcal{F}}\mathbb{o}t$\\
      Grounding time complexity & $t{C_{w}}^{t\mathbb{\Lambda}}\mathbb{Z}^{C_\Lambda}$ & $t{C_w}^{\mathbb{\Lambda}}\mathbb{Z}^{C_\Lambda}$ & ${C_w}^{\mathbb{\Lambda}}\mathbb{Z}^{C_\Lambda}$\\ 
    \end{tabular}
  \end{footnotesize}
  \label{tab:complexity}
  \end{center}
  \vspace{-4ex}
\end{table*}

Incremental estimation in temporal grounding graphs relies on  
propagating a state $\KB{t}$ while retaining 
visual observations $\VisualObservations{0:t}$ 
trading off the runtime of grounding a single utterance with the 
space and time complexity of estimating perceptual groundings from visual observations.
Table~\ref{tab:complexity} provides a complexity analysis for the proposed approach 
compared to two common alternatives:
the first column of the table presents the analysis corresponding to a model
that relies entirely on the observation history without any state maintenance,
i.e., estimates
$P(\Correspondences{t} | \Language{0:t}, \VisualObservations{0:t}, \Grounding{t} )$.
The last column corresponds to a model that maintains a \emph{full} symbolic state
without retaining either the visual or linguistic observations. 
We introduce the following notation for the analysis: $\mathbb{\Lambda}$ and
$\mathbb{Z}$ are the worst case longest sentence and video requiring the most
number of detections, $C_w$ is the declarative symbol with the largest state space,
$C_\Lambda$ is the number of participants in the event described by the
worst case sentence, $\mathbb{\Gamma}^{\mathcal{F}}$ and
$\mathbb{\Gamma}^{\mathcal{P}}$ are the number of the factual and perceptual
grounding predicates, and $\mathbb{o}$ is the number of possible object
instances.

Not keeping any state results in low space complexity
$O(\mathbb{\Lambda}t+\mathbb{Z}t)$, while having high grounding time complexity
$O(t{C_{w}}^{t\mathbb{\Lambda}}\mathbb{Z}^{C_\Lambda})$.
Note the exponential dependency on $t$, the number of time steps.
Reasoning about any new sentence requires re-reasoning about all previously seen
sentences and any correlations between those sentences resulting in an
exponentially increasing joint distribution.
Even for short exchanges this runtime is infeasible.
Conversely, keeping all perceptual state in a symbolic manner results in
extremely efficient inference of groundings,
$O({C_w}^{\mathbb{\Lambda}}\mathbb{Z}^{C_\Lambda})$.
The likelihoods of any declarative groundings are already recorded and only
groundings which are relevant to the given command must be updated from time
$t-1$ to $t$.
Yet this process must record all possible inferences for any grounding in any
previous observation so they can be available for grounding when the stimulus is
discarded.
Doing so is prohibitively expensive,
$O(\mathbb{\Gamma}^{\mathcal{P}}\mathbb{o}^{C_{\Lambda}C_{w}}t+\mathbb{\Gamma}^{\mathcal{F}}\mathbb{o}t)$,
due to the arity of groundings; even a binary grounding requires storing a fact
about every pair of possible objects.

We strike a middle ground between these alternatives by taking advantage of two
facts.
First, any one sentence requires a small number of declarative groundings which
can be estimated using prior visual observations in time linear in the size of
those observations.
Second, storing factual groundings removes the need to re-run inference over
prior utterances dramatically speeding up grounding time to
$O(t{C_w}^{\mathbb{\Lambda}}\mathbb{Z}^{C_\Lambda})$.
Compared to not having any stored state, we see significant speedup due to the
the lack of $t$ as an exponent while adding only a small amount of storage,
$O(\mathbb{\Gamma}^{\mathcal{F}}\mathbb{o}t)$ which increases by at most a small
constant factor with each new sentence,
Note that the propagated state contains only factual groundings. We revert to using the visual
observations to save on exponential increase in storage complexity and record
facts to lower the exponent of inferring groundings.

\section{Evaluation}
\label{sec:evaluation}
The system was deployed on the Baxter Research Robot operating 
on a tabletop workspace. 
The robot observed the workspace using images captured using a cross-calibrated 
Kinect version 2 RGB-D sensor at $\sim$20Hz with 1080x760 resolution.
Objects were localised using a multi-scale sub-window search in image space with
filtering using depth information.
A binary SVM with colour histogram features was used for object recognition.
The robot engaged in several interactions with human agents that were speaking
or typing natural language sentences providing information, narrating the
events, or requesting the execution of a command.
Spoken commands from the human operator were converted to text using an Amazon Echo Dot.
We demonstrate the space of capabilities of the model through a qualitative
evaluation and create a corpus of interactions to demonstrate its
robustness.\footnote{%
  Video demonstrations and the corpus used for quantitative evaluation are
  available at: \url{http://toyota.csail.mit.edu/node/28}}
The model was evaluated qualitatively and quantitatively, both of which we discuss next. 
%

\subsection{Qualitative Results}
\begin{figure}[h!]
  \captionsetup[subfigure]{width=1.0\columnwidth}
  \centering
    \subfloat [][ \scriptsize{``The cracker box on the table is my snack." ``Pick up my snack.''\\
    Inference disambiguates the object for the pick action by relying on the updated fact that the box is the agent's snack.}] {
    \includegraphics[width=0.81\columnwidth]{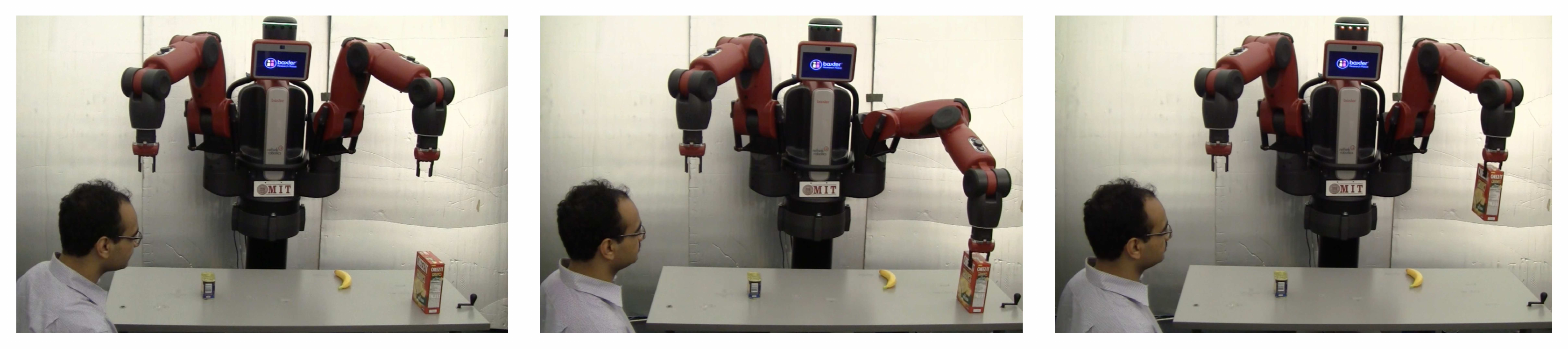} \label{a} }\vspace{-2ex}

    \subfloat [][\scriptsize{``Lift the box that I put down.''\\
    The object to be lifted is disambiguated in the context of a video depicting an action.}] 
    {\includegraphics[width=0.81\columnwidth]{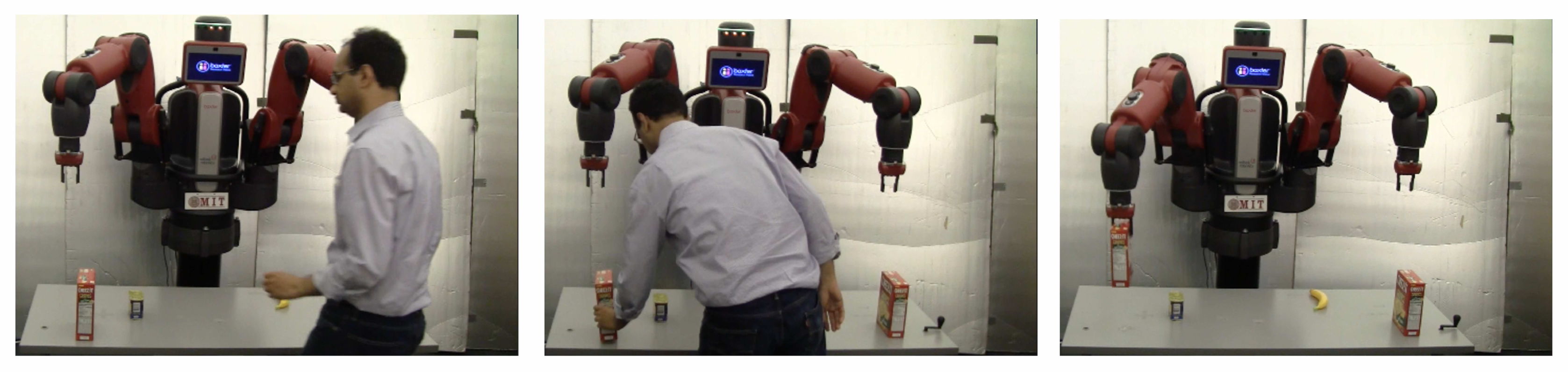} \label{b} }\vspace{-2ex}

    \subfloat [][\scriptsize{``The box and the can are my snack.'' ``Pack up my snack.''\\
    The inferred grounding is an abstract aggregation (\emph{snack}) composed of two tracks 
    corresponding to the can and the box resulting in a multi-action task.}] 
    {\includegraphics[width=0.81\columnwidth]{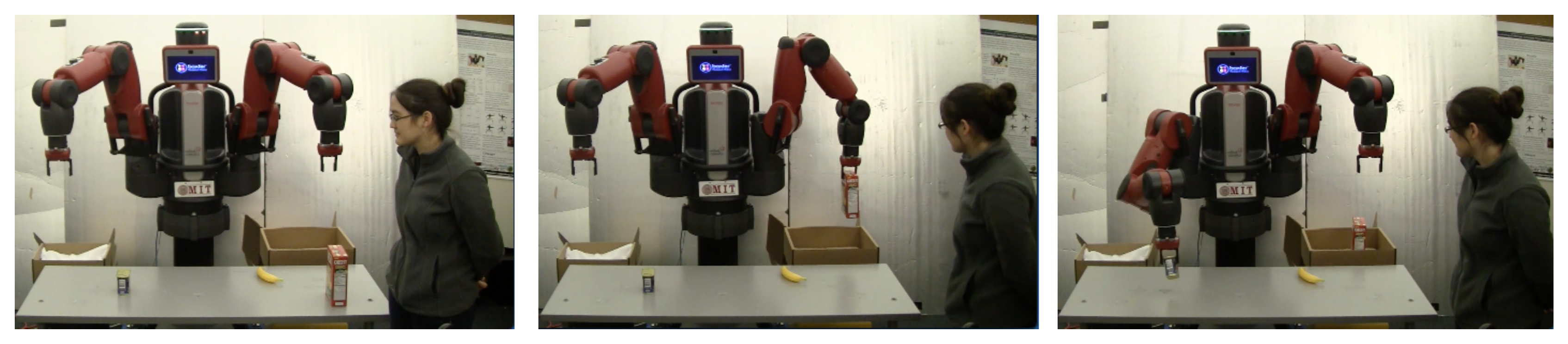} \label{c} }\vspace{-2ex}
    
    \subfloat [][\scriptsize{``The box I will put down is my snack.'' ``Pick it up.''\\
    A combination of syntactic and visual features are used to resolve the co-reference.}] 
    {\includegraphics[width=0.81\columnwidth]{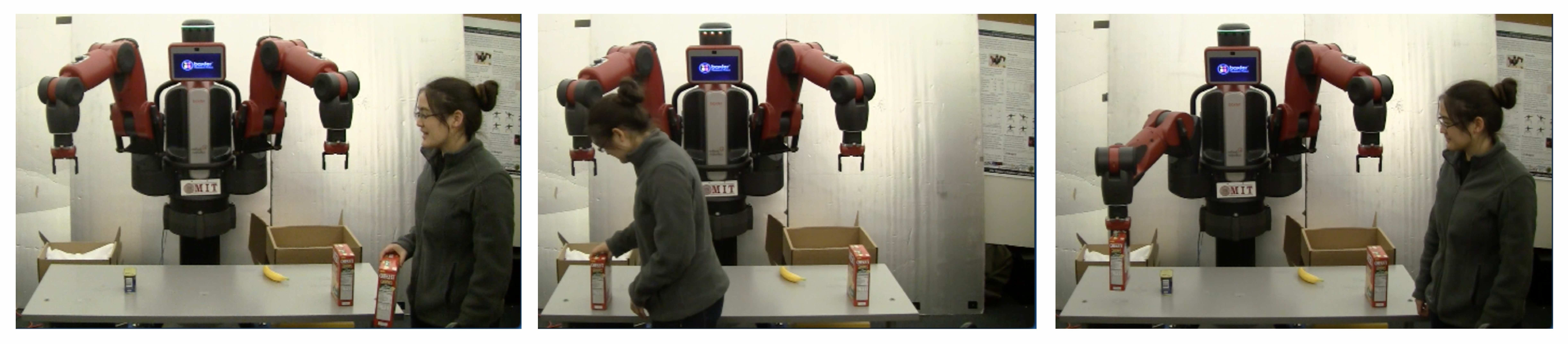} \label{d} }\vspace{-2ex}
    
    \subfloat [][\scriptsize{``The fruit on the table is mine.''
      ``The green fruit is mine.'' ``Point at my fruit.''\\
    Partial information from ambiguous statements is fused together to select an object.}] 
    {\includegraphics[width=0.81\columnwidth]{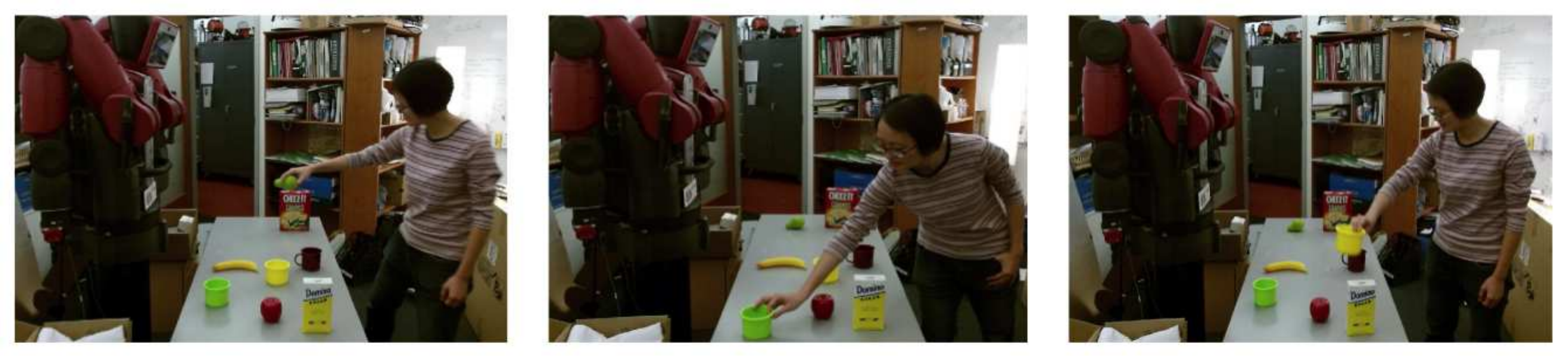} \label{e} }
    
    \caption{Examples of grounding instructions given visual observations and
      factual information gathered through one or more natural-language
      interactions.}
   \label{figure-qualitative-results}
\end{figure}
Figure \ref{figure-qualitative-results} presents representative examples of grounding a sequence of instructions 
from the human operator using the proposed model. 
%
A robot performs five tasks requiring a combination of state keeping,
disambiguating partial information, and observations of its environment.
At times, such as in example (d), all such capabilities are needed as part of a
joint inference process in order to arrive at the correct grounding.
As the scenario unfolds the robot becomes more certain about its groundings,
eventually being able to perform its assigned task.

\subsection{Quantitative Evaluation}

To evaluate our approach quantitatively we collected a video corpus of humans
performing actions while providing declarative facts and commands for the robot
to execute.
Our corpus consists of longer videos composed by combining 96 short, 3 second
long, videos consisting of a person performing one action out of 5 (\emph{pick
  up}, \emph{put down}, \emph{slide}, \emph{move toward}, \emph{move away from})
with one of eight objects from the YCB data set \cite{calli2015ycb} (3 fruit, 3
cups, and 2 boxes) where each object had one of three colors (red, green,
yellow), and was either small or large, and could be on the table or on top of
another object.
The 96 short videos were collected by filming users, not the authors,
interacting with objects on a tabletop.
They were given generic directions such as ``slide a cup on the table".
The cues ensured all possible pairings of actions, objects and spatial layouts.
These shorter videos were stitched together to form the final video corpus
consisting of between one and three seed videos concatenated together, an
optional declarative sentence associated with that smaller component video, and
a final command for the robot to perform an action with one of the objects.
Out of the possible 96$^3$ videos we created 255 video-sentence pairs.
Of these videos, 180 depicted either one or two actions performed by a human
followed by a command whose interpretation refers to one or both actions.
For example, ``pick up the red object the person put down'' associated with a
clip where one object was picked up and another was put down making the correct
interpretation of this command depend on the actions observed.
The remainder of the corpus, 75 videos, depicts either two or three actions
performed by a human with optional associated declarative facts that may refer
to those actions along with a final command.
For example, a sequence of videos is paired with the sentences ``the green object
is the oldest'', ``the fruit on the table is the oldest'', followed by ``point
at the oldest object''.
Sentential annotations for the quantitative evaluation were provided by the
authors but in the future we intend to test on more diverse user-generated
utterances.

A human judge watched these video sequences captioned with the sentences they
were paired with.
That judge annotated the expected action the robot should perform and the target
object with which it should be performed.
An inferred robotic command was only considered correct when it performed the
correct action on the correct objects in the intended location.
These annotations were compared with those generated automatically resulting in
an accuracy of 92.5\% demonstrating the effectiveness of the state propagation 
in the model.
Chance performance is 
$\frac{1}{27}$, corresponding to a  $\frac{1}{9}$
chance of choosing the correct object and a $\frac{1}{3}$ chance of performing
the correct action with that object.
Performance on short videos was worse, with 90.2\% of inferences being correct,
than on long videos, where 94.7\% of inferred actions and target objects were
correct.
While longer videos provide more opportunities for failure they also provide
more context for inference.
Failures occurred largely due to errors in perception.
Actions which moved objects toward or away from the camera were difficult to
perceive.
Several objects were occluded while an action was performed which occasionally
led to an incorrect interpretation of that action.

\section{Conclusion}
\label{sec:conclusions}

The model presented here significantly extends the space of commands that robots
can understand.
It incorporates factual knowledge from prior linguistic interactions and visual
observations of a workspace including the actions of other agents in that
workspace into a coherent approach that performs joint inference to understand a
sentence providing a command or new factual knowledge.
Knowledge accrues and is refined over time through further linguistic
interactions and observations.
We intend to extend this model to ground to sequences of actions and collections
of objects, to engage in dialog while executing multi-step actions, to keep
track of and infer the locations of partially observed objects, and to serve as
a basis for a grounded and embodied model of language acquisition.

\section*{Acknowledgements}
\begin{footnotesize}
We acknowledge funding support in part by the Toyota Research Institute Award Number LP-C000765-SR;
the Robotics Collaborative Technology Alliance (RCTA) of the US Army;
the Center for Brains, Minds, and Machines (CBMM) funded by NSF STC award CCF-1231216; 
and AFRL contract No. FA8750-15-C-0010.
We thank Naomi Schurr and Yen-Ling Kuo for assistance during system evaluation.
\end{footnotesize}


\pagebreak
\begin{small}
\bibliographystyle{named}
\bibliography{rtmlv}

\begin{thebibliography}{}

\bibitem[\protect\citeauthoryear{Andreas and
  Klein}{2015}]{andreas2015alignment}
Jacob Andreas and Dan Klein.
\newblock Alignment-based compositional semantics for instruction following.
\newblock In {\em Empirical Methods in Natural Language Processing (EMNLP)},
  2015.

\bibitem[\protect\citeauthoryear{Artzi and Zettlemoyer}{2013}]{artzi2013weakly}
Yoav Artzi and Luke Zettlemoyer.
\newblock Weakly supervised learning of semantic parsers for mapping
  instructions to actions.
\newblock {\em Transactions of the Association for Computational Linguistics},
  1:49--62, 2013.

\bibitem[\protect\citeauthoryear{Barrett \bgroup \em et~al.\egroup
  }{2016}]{barbu2013saying}
D.~P. Barrett, A.~Barbu, N.~Siddharth, and J.~M. Siskind.
\newblock Saying what you're looking for: Linguistics meets video search.
\newblock {\em IEEE Transactions on Pattern Analysis and Machine Intelligence},
  38(10):2069--2081, Oct 2016.

\bibitem[\protect\citeauthoryear{Berant \bgroup \em et~al.\egroup
  }{2013}]{berant2013freebase}
J.~Berant, A.~Chou, R.~Frostig, and P.~Liang.
\newblock Semantic parsing on {F}reebase from question-answer pairs.
\newblock In {\em Empirical Methods in Natural Language Processing (EMNLP)},
  2013.

\bibitem[\protect\citeauthoryear{Berzak \bgroup \em et~al.\egroup
  }{2016}]{berzak2016you}
Yevgeni Berzak, Andrei Barbu, Daniel Harari, Boris Katz, and Shimon Ullman.
\newblock Do you see what {I} mean? {V}isual resolution of linguistic
  ambiguities.
\newblock {\em arXiv preprint arXiv:1603.08079}, 2016.

\bibitem[\protect\citeauthoryear{Calli \bgroup \em et~al.\egroup
  }{2015}]{calli2015ycb}
Berk Calli, Arjun Singh, Aaron Walsman, Siddhartha Srinivasa, Pieter Abbeel,
  and Aaron~M Dollar.
\newblock The {YCB} object and model set: Towards common benchmarks for
  manipulation research.
\newblock In {\em International Conference on Advanced Robotics (ICAR)}, pages
  510--517. IEEE, 2015.

\bibitem[\protect\citeauthoryear{Cantrell \bgroup \em et~al.\egroup
  }{2010}]{cantrell2010robust}
Rehj Cantrell, Matthias Scheutz, Paul Schermerhorn, and Xuan Wu.
\newblock Robust spoken instruction understanding for {HRI}.
\newblock In {\em Proceedings of the 5th ACM/IEEE international conference on
  Human-robot interaction}, pages 275--282. IEEE Press, 2010.

\bibitem[\protect\citeauthoryear{Chen and Mooney}{2011}]{chen2011learning}
David~L Chen and Raymond~J Mooney.
\newblock Learning to interpret natural language navigation instructions from
  observations.
\newblock In {\em AAAI}, volume~2, pages 1--2, 2011.

\bibitem[\protect\citeauthoryear{Chen \bgroup \em et~al.\egroup
  }{2010}]{chen2010training}
David~L Chen, Joohyun Kim, and Raymond~J Mooney.
\newblock Training a multilingual sportscaster: Using perceptual context to
  learn language.
\newblock {\em Journal of Artificial Intelligence Research}, pages 397--435,
  2010.

\bibitem[\protect\citeauthoryear{Chung \bgroup \em et~al.\egroup
  }{2015}]{chung2015performance}
Istvan Chung, Oron Propp, Matthew~R Walter, and Thomas~M Howard.
\newblock On the performance of hierarchical distributed correspondence graphs
  for efficient symbol grounding of robot instructions.
\newblock In {\em IEEE/RSJ International Conference on Intelligent Robots and
  Systems (IROS)}, pages 5247--5252. IEEE, 2015.

\bibitem[\protect\citeauthoryear{Ghahramani \bgroup \em et~al.\egroup
  }{1997}]{ghahramani1997factorial}
Zoubin Ghahramani, Michael~I Jordan, and Padhraic Smyth.
\newblock Factorial hidden {M}arkov models.
\newblock {\em Machine learning}, 29(2-3):245--273, 1997.

\bibitem[\protect\citeauthoryear{Guadarrama \bgroup \em et~al.\egroup
  }{2013}]{guadarrama2013grounding}
Sergio Guadarrama, Lorenzo Riano, Dave Golland, Daniel Gouhring, Yangqing Jia,
  David Klein, Pieter Abbeel, and Trevor Darrell.
\newblock Grounding spatial relations for human-robot interaction.
\newblock In {\em International Conference on Intelligent Robots and Systems
  (IROS)}, pages 1640--1647. IEEE, 2013.

\bibitem[\protect\citeauthoryear{Hemachandra \bgroup \em et~al.\egroup
  }{2015}]{hemachandra15}
Sachithra Hemachandra, Felix Duvallet, Thomas~M. Howard, Nicholas Roy, Anthony
  Stentz, and Matthew~R Walter.
\newblock Learning models for following natural language directions in unknown
  environments.
\newblock In {\em Robotics and Automation (ICRA), 2015 IEEE International
  Conference on}, pages 5608--5615. IEEE, 2015.

\bibitem[\protect\citeauthoryear{Howard \bgroup \em et~al.\egroup
  }{2014}]{howard2014efficient}
Thomas~M Howard, Istvan Chung, Oron Propp, Matthew~R Walter, and Nicholas Roy.
\newblock Efficient natural language interfaces for assistive robots.
\newblock In {\em IEEE/RSJ Int’l Conf. on Intelligent Robots and Systems
  (IROS) Work. on Rehabilitation and Assistive Robotics}, 2014.

\bibitem[\protect\citeauthoryear{Katz}{1988}]{katz1988using}
Boris Katz.
\newblock Using {E}nglish for indexing and retrieving.
\newblock In {\em Recherche d'Information Assist{\'e}e par Ordinateur (RIAO)},
  pages 314--332, 1988.

\bibitem[\protect\citeauthoryear{Kim and Mooney}{2012}]{kim2012unsupervised}
Joohyun Kim and Raymond~J Mooney.
\newblock Unsupervised {PCFG} induction for grounded language learning with
  highly ambiguous supervision.
\newblock In {\em Proceedings of the Joint Conference on Empirical Methods in
  Natural Language Processing (EMNLP)}, pages 433--444. Association for
  Computational Linguistics, 2012.

\bibitem[\protect\citeauthoryear{Liu \bgroup \em et~al.\egroup
  }{2016}]{liu2016taskstructure}
C.~Liu, S.~Yang1, S.~Saba-Sadiya1, N.~Shukla, Y.~He, Song-Chun Zhu, , and J.~Y.
  Chai.
\newblock Jointly learning grounded task structures from language instruction
  and visual demonstration.
\newblock In {\em Empirical Methods in Natural Language Processing (EMNLP)},
  2016.

\bibitem[\protect\citeauthoryear{Matuszek \bgroup \em et~al.\egroup
  }{2014}]{MatuszekAAAI2014}
Cynthia Matuszek, Liefeng Bo, Luke Zetllemoyer, and Dieter Fox.
\newblock Learning from unscripted deictic gesture and language for human-robot
  interactions.
\newblock In {\em Proceedings of the National Conference on Artificial
  Intelligence (AAAI)}, Qu{\'e}bec City, Quebec, Canada, March 2014.

\bibitem[\protect\citeauthoryear{Misra \bgroup \em et~al.\egroup
  }{2016}]{misra2016tell}
Dipendra~K Misra, Jaeyong Sung, Kevin Lee, and Ashutosh Saxena.
\newblock Tell me {D}ave: Context-sensitive grounding of natural language to
  manipulation instructions.
\newblock {\em The International Journal of Robotics Research},
  35(1-3):281--300, 2016.

\bibitem[\protect\citeauthoryear{Paul \bgroup \em et~al.\egroup
  }{2016}]{Paul-RSS-16}
Rohan Paul, Jacob Arkin, Nicholas Roy, and Thomas~M. Howard.
\newblock Efficient grounding of abstract spatial concepts for natural language
  interaction with robot manipulators.
\newblock In {\em Proceedings of Robotics: Science and Systems}, Ann Arbor,
  Michigan, June 2016.

\bibitem[\protect\citeauthoryear{Russell and Norvig}{1995}]{russell1995modern}
Stuart Russell and Peter Norvig.
\newblock Artificial intelligence: A modern approach.
\newblock {\em Prentice-Hall, Egnlewood Cliffs}, 25:27, 1995.

\bibitem[\protect\citeauthoryear{Tellex \bgroup \em et~al.\egroup
  }{2011}]{tellex2011understanding}
Stefanie Tellex, Thomas Kollar, Steven Dickerson, Matthew~R. Walter,
  Ashis~Gopal Banerjee, Seth~J Teller, and Nicholas Roy.
\newblock Understanding natural language commands for robotic navigation and
  mobile manipulation.
\newblock In {\em Proceedings of the National Conference on Artificial
  Intelligence (AAAI)}, 2011.

\bibitem[\protect\citeauthoryear{Walter \bgroup \em et~al.\egroup
  }{2014}]{walter14a}
Matthew~R Walter, Sachithra Hemachandra, Bianca Homberg, Stefanie Tellex, and
  Seth Teller.
\newblock A framework for learning semantic maps from grounded natural language
  descriptions.
\newblock {\em The International Journal of Robotics Research (IJRR)},
  33(9):1167--1190, August 2014.

\bibitem[\protect\citeauthoryear{Yu \bgroup \em et~al.\egroup
  }{2015}]{yu2015compositional}
Haonan Yu, N~Siddharth, Andrei Barbu, and Jeffrey~Mark Siskind.
\newblock A compositional framework for grounding language inference,
  generation, and acquisition in video.
\newblock {\em J. Artif. Intell. Res. (JAIR)}, 52:601--713, 2015.

\bibitem[\protect\citeauthoryear{Zettlemoyer and
  Collins}{2007}]{zettlemoyer2007online}
Luke~S Zettlemoyer and Michael Collins.
\newblock Online learning of relaxed {CCG} grammars for parsing to logical
  form.
\newblock In {\em Proceedings of Empirical Methods in Natural Language
  Processing and Computational Nautal Language Learning (EMNLP-CoNLL)}, pages
  678--687, Prague, 2007.

\end{thebibliography}
\end{small}

\end{document}